%% file: main.tex
\documentclass[journal]{IEEEtran}
\usepackage{amsmath,amsfonts}
\usepackage{amsthm,amssymb}
\usepackage{mathrsfs}
\usepackage{CJKutf8}

\usepackage{algorithmic}
\usepackage{algorithm}
\usepackage{array}
\usepackage[caption=false,font=normalsize,labelfont=sf,textfont=sf]{subfig}
\usepackage{textcomp}
\usepackage{stfloats}
\usepackage{url}
\usepackage{verbatim}
\usepackage{graphicx}
\usepackage{cite}

\usepackage{soul}
\usepackage{bm}
\usepackage{booktabs}
\usepackage{txfonts}
\usepackage{booktabs}
\usepackage{multirow}
\usepackage{caption3}
\usepackage{xcolor}
\usepackage{pifont}
\usepackage{threeparttable}

\usepackage{hyperref}
\usepackage{float}

\usepackage{textcomp}
\usepackage{stfloats}
\usepackage{url}
\usepackage{verbatim}
\usepackage{graphicx}

\usepackage{cite}

\usepackage{soul}
\usepackage{bm}
\usepackage{booktabs}
\usepackage{txfonts}
\usepackage{booktabs}
\usepackage{multirow}
\usepackage{caption3}
\usepackage{caption}
\usepackage{xcolor}
\usepackage{float}
\usepackage[table]{xcolor} 
\definecolor{darkgray}{rgb}{0.7, 0.7, 0.7} 
\definecolor{gray}{rgb}{0.8, 0.8, 0.8}     
\definecolor{lightgray}{rgb}{0.9, 0.9, 0.9} 

\usepackage{pifont}
\usepackage{threeparttable}

\usepackage{array}
\usepackage{multirow}

\newcolumntype{L}[1]{>{\raggedright\arraybackslash}p{#1}}
\newcolumntype{C}[1]{>{\centering\arraybackslash}p{#1}}

\begin{document}
\begin{CJK*}{UTF8}{gbsn}
\title{SFR-Net: Learning Scale-Frustum Representations for Ultra-Wide Area Remote Sensing \\ Image Segmentation}

\author{Chuyu~Zhong, Keyan~Chen,~Qinzhe~Yang,~Bowen~Chen\\Zhengxia~Zou,~\IEEEmembership{Senior~Member,~IEEE},~and~Zhenwei~Shi$^\star$,~\IEEEmembership{Senior~Member,~IEEE}

\thanks{This work was supported in part by the National Natural Science Foundation of China under Grants U24B20177, 62125102, U25A20401 and 62471014, in part by the Inner Mongolia Autonomous Region Science and Technology Planning Project under Grant 2025YFHH0124, and in part by the Fundamental Research Funds for the Central Universities. \emph{(Corresponding Author: Zhenwei Shi (shizhenwei@buaa.edu.cn))}}

\thanks{Chuyu Zhong, Bowen Chen, Zhengxia Zou, and Zhenwei Shi are with the Department of Aerospace Intelligent Science and Technology, School of Astronautics, Beihang University, Beijing 100191, China, and also with the Key Laboratory of Spacecraft Design Optimization and Dynamic Simulation Technologies, Ministry of Education, Beihang University, Beijing 100191, China. Qinzhe Yang is with Shen Yuan Honors College, Beihang University, Beijing 100191, China. Keyan Chen is with the College of Computing and Data Science, Nanyang Technological University, Singapore. 

}
}

\maketitle


\input{sec/abstract}

\input{sec/introduction}

\input{sec/related_works}

\input{sec/methodology}

\input{sec/experiments_and_analysis}

\input{sec/conclusion}

\input{sec/biography}

\end{CJK*}
\end{document}

%% file: sec/abstract.tex
\begin{abstract}

Pixel count and geographical coverage are two key characteristics of remote sensing images. Existing remote sensing image segmentation methods typically focus on images with either a small pixel count or a large pixel count but limited geographical coverage. In this paper, we introduce a novel segmentation task targeting ultra-wide area (UWA) remote sensing images, characterized by both a large pixel count and extremely wide geographical coverage. The core challenges of UWA segmentation lie in simultaneously handling ground objects with significantly varying scales and maintaining long-range contextual semantic continuity. To address these challenges, we propose the Scale-Frustum Representation Network (SFR-Net). Inspired by the viewing frustums of remote sensing images captured from different altitudes, we construct scale-frustum representations, enabling unified modeling of ground objects and contextual features at different scales. Furthermore, we design a cascaded cross-scale fusion mechanism to effectively integrate these representations, enhancing local semantic understanding while ensuring long-range contextual continuity. Experimental results on GID and FBPS demonstrate that SFR-Net achieves state-of-the-art performance, improving mIoU by 1.72\% and 4.29\%, respectively, over the strongest competing methods. In addition, the proposed scale-frustum representations can be integrated into generic segmentation networks to improve both segmentation accuracy and convergence speed. The implementation code will be publicly available at \url{https://github.com/ChuyuZhong/SFR-Net}.

\end{abstract}


%% file: sec/introduction.tex
\section{Introduction}

%
%
%
%

 

\begin{figure}[t]
  \centering
   \includegraphics[width=1\linewidth]{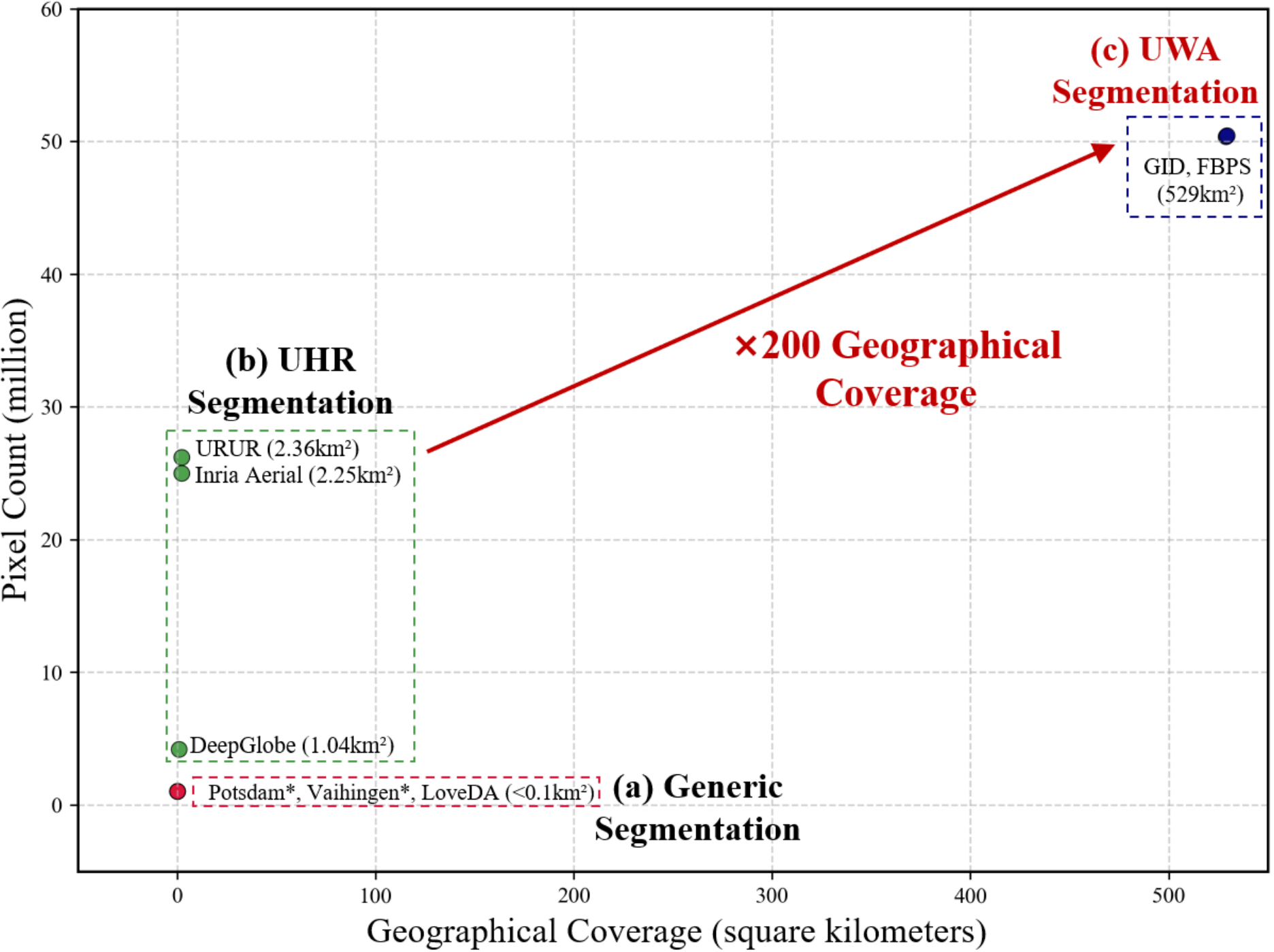}
   \caption{Defining the ultra-wide area (UWA) segmentation task. This plot compares (a) Generic, (b) UHR, and (c) UWA segmentation in terms of pixel count and geographical coverage. The ``*'' denotes that original dataset images were cropped to sizes commonly used in research.}
   \label{fig:uwa}
\end{figure}

\begin{figure}[t]
  \centering
   \includegraphics[width=1\linewidth]{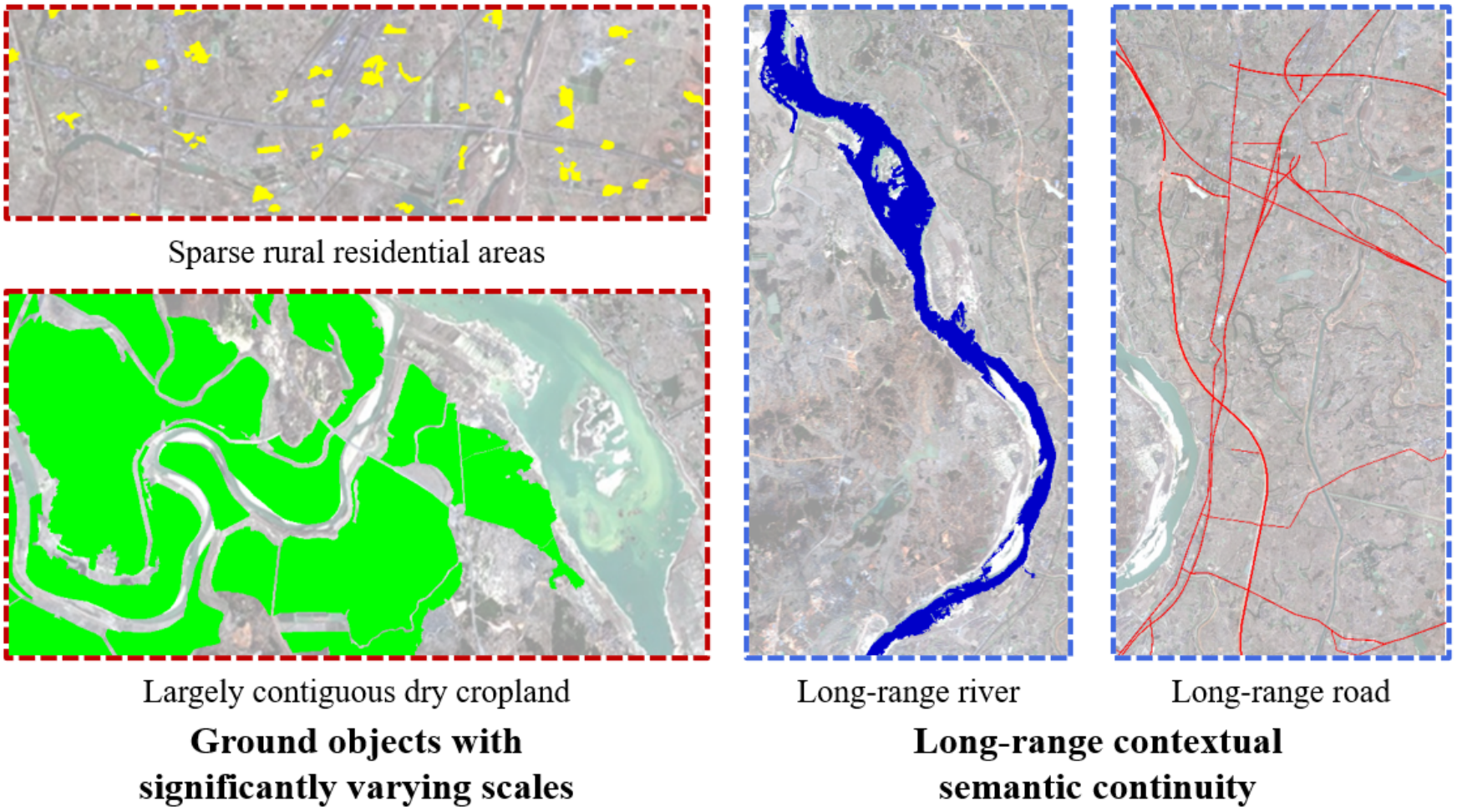}

   \caption{Visual examples of the two core challenges in UWA segmentation. Left: The challenge of handling significantly varying scales, illustrated by contrasting small, sparse buildings with large, contiguous cropland. Right: The challenge of maintaining long-range contextual continuity, exemplified by rivers and roads.}
   \label{fig:uwa_challenge}
\end{figure}

\IEEEPARstart{P}ixel count and geographical coverage are two key characteristics of remote sensing images. As shown in Fig.~\ref{fig:uwa}, ultra-wide area (UWA) remote sensing images refer to those with a large pixel count (e.g., $\ge 5000 \times 5000$) and extremely wide geographical coverage (e.g., $\ge 500$ km$^2$). Semantic segmentation of UWA remote sensing images can not only provide refined details about ground objects but also facilitate the semantic understanding of large-scale scenes. Although representative datasets \cite{gid,fbps} have been used in previous studies, they were usually treated as general land-cover segmentation benchmarks or processed with cropping-based pipelines.In this paper, we explicitly formulates UWA segmentation as a task setting that jointly emphasizes large pixel count, extremely wide geographical coverage, complex land-cover composition, and long-range semantic continuity. As shown in Fig.~\ref{fig:uwa_challenge}, unlike images with a small pixel count or limited geographical coverage, UWA remote sensing images simultaneously present ground objects with significantly varying scales and demand long-range contextual semantic continuity. These characteristics collectively make this task extremely challenging.

Generic semantic segmentation \cite{fcn,unet,segnet,maskformer,sam,rsprompter} typically targets images with a small pixel count, as shown in Fig.~\ref{fig:uwa}. Due to computational constraints, these methods are unable to directly process entire UWA remote sensing images. A common strategy is to split the UWA image into smaller, manageable patches. While this processing strategy can capture abundant local details, it often under-utilizes long-range context, impeding the maintenance of long-range semantic continuity. Additionally, these methods generalize poorly to ground objects with significantly varying scales, severely limiting their performance in UWA segmentation. Some improved generic semantic segmentation methods \cite{fctl,rest} can incorporate contextual information; however, they demand significant computational resources.

In recent years, methods targeting ultra-high resolution (UHR) image segmentation \cite{glnet,grnet,uhrsnet,isdnet,gpwformer} have also been proposed, which directly process images with a large pixel count, as shown in Fig.~\ref{fig:uwa}. To balance speed and accuracy, these methods typically adopt a lightweight encoder to directly process the entire UHR image. While this strategy achieves excellent segmentation performance on images with limited geographical coverage, it encounters inherent challenges when applied to UWA remote sensing images. Specifically, they have difficulty representing refined land cover details. Consequently, these methods struggle to handle small targets and multi-category ground objects in UWA remote sensing images, which limits their applicability to UWA segmentation.

The enormous geographical coverage of UWA images inherently encompasses continuous scales, from refined local details to global macroscopic overviews. This unique characteristic is similar to the viewing frustum of sensors at different altitudes during the imaging process. Inspired by this similarity, we present the Scale-Frustum Representation Network (SFR-Net). To achieve the perception of ground objects at different scales, we construct scale-frustum representations (SFR) for ultra-wide area remote sensing images. This representation can unify the modeling of local observation, short-range observation, and long-range observation, providing rich contextual features. We also propose scale embeddings to enable the network to better perceive contextual information at different scales. Furthermore, to model long-range contextual semantic continuity, we design a cascaded cross-scale fusion (CCSF) module that gradually injects short-range and long-range contextual features into local features.
Experimental results demonstrate that our method achieves state-of-the-art performance on UWA segmentation tasks. Additionally, experiments show that the proposed scale-frustum representations can be integrated into generic segmentation networks, leading to improved segmentation accuracy and faster convergence.

Overall, the contributions of this paper can be summarized as follows:

1) We propose the Scale-Frustum Representation Network (SFR-Net), which unifies the modeling of ground objects and contextual features at different scales.

2) We introduce a cascaded cross-scale fusion (CCSF) module, which significantly enhances the long-range semantic continuity of segmentation results.

3) We conduct extensive experiments demonstrating that our method achieves state-of-the-art performance on the UWA task, and that our SFR can significantly improve the performance and convergence speed of generic segmentation methods.

%% file: sec/related_works.tex
\section{Related Works}

This section reviews three key areas relevant to our proposed method: generic semantic segmentation, ultra-high resolution segmentation, and whole-scene remote sensing image segmentation.

\subsection{Generic Semantic Segmentation}

Since FCN \cite{fcn} formally proposed the task of semantic segmentation, a multitude of networks and benchmarks were introduced. U-Net \cite{unet} applied an encoder-decoder architecture to the semantic segmentation task, effectively extracting and utilizing features across different scales. Such a foundational design has been widely adopted in subsequent models. PSPNet \cite{pspnet} introduced the pyramid pooling module, enhancing the model's ability to capture multi-scale features. Similarly, ICNet \cite{icnet} employed cascaded image inputs and cascaded label guidance strategies to efficiently fuse multi-level features, achieving superior segmentation efficiency. The DeepLab series \cite{deeplabv3plus} introduced atrous convolution and conditional random fields to further improve segmentation accuracy. D-LinkNet \cite{dlinknet} designed an improved encoder-decoder architecture combined with a ResNet \cite{resnet} backbone. MSCA \cite{msca} integrated features of varying dimensions and depths, conducting feature fusion across both long-range and local contexts. In addition, some methods focused on improving both accuracy and speed. BiSeNet \cite{bisenet} originally proposed the concept of a two-stream network to achieve high-precision real-time semantic segmentation. STDC \cite{stdc} readjusted its two-stream architecture, using a simple shortcut connection and a fusion mechanism to introduce low-level feature information.

With the advent of the Transformer architecture \cite{vit}, several works \cite{segformer,swin,maskformer,mask2former,sam,rsprompter} also explored Transformer-based semantic segmentation models. ViT \cite{vit} first introduced the Transformer architecture into computer vision. By dividing images into several patches, it converts images into sequential inputs and utilizes the self-attention mechanism to capture global features, achieving strong performance on image classification. SETR \cite{setr} further applied the Transformer architecture to semantic segmentation tasks, demonstrating that the plain ViT backbone also possesses strong advantages in semantic segmentation. SegFormer \cite{segformer} optimized the performance of the plain ViT backbone in semantic segmentation tasks by introducing a CNN-like pooling step to fuse multi-level features. Swin Transformer \cite{swin} further introduced a local window attention mechanism, effectively reducing the computational complexity of Transformer-based models while maintaining the ability to model long-range dependencies. Similar to BiSeNet, DANet \cite{danet} introduced dual-stream attention, which can separately capture semantic dependencies along the spatial and channel dimensions. Similarly, SCAttNet \cite{scattnet} constructed a spatial attention module and a channel attention module, thereby improving semantic segmentation for high-resolution remote sensing images. LANet \cite{lanet} integrated features extracted by low-level CNNs into high-level CNN features through an attention mechanism, obtaining enhanced high-level and low-level features for fused decoding, thereby achieving high segmentation accuracy. SAM \cite{sam} built a data engine, utilizing a massive amount of auto-annotated and human-assisted annotated image segmentation data to train a visual foundation model that can be directly transferred to various segmentation tasks. Building upon this, RSPrompter \cite{rsprompter} constructed a prompt learning framework for remote sensing images based on the SAM foundation model, adaptively generating prompt inputs to make it applicable to remote sensing instance segmentation tasks. In addition to visual prompts, language-guided segmentation has also been explored in remote sensing \cite{rsrefseg,taco}. Specifically, MaskFormer \cite{maskformer,mask2former} adopted a unified architecture, enabling simultaneous semantic segmentation and instance segmentation.

Recently, the Mamba \cite{mamba} architecture has emerged as a novel sequence processing model. By leveraging the advantages of state space models \cite{ssm}, it effectively processes long sequences with linear time complexity, prompting numerous studies to adapt it for the vision domain \cite{dynamicvis}. RS-Mamba \cite{rsmamba}, a variant tailored for remote sensing semantic segmentation, introduces a diagonal image scanning path to enhance feature representation in global modeling. RS3Mamba \cite{rs3mamba}, a dual-branch network designed for remote sensing tasks, integrates visual state space blocks and a collaborative completion module to effectively fuse global and local features, thereby significantly improving segmentation accuracy. Similarly, UNetMamba \cite{unetmamba} achieves efficient and lightweight semantic segmentation.

Although the above methods have achieved strong performance on semantic segmentation tasks for remote sensing images with a small pixel count, they still struggle to directly handle UWA segmentation due to computational constraints. A common strategy is to divide a UWA image into smaller and more manageable patches, which inevitably leads to fragmented contextual semantics. Another more straightforward solution is to downsample the entire UWA image, but such a naive practice inevitably sacrifices a large amount of refined local detail. In this paper, we demonstrate that our proposed scale-frustum representations can greatly enhance the performance of these methods with fewer training iterations.

\subsection{Ultra-high Resolution Segmentation}

In recent years, several semantic segmentation methods specifically designed for ultra-high resolution remote sensing images have gradually emerged, which can be broadly categorized into three groups: global-local branch fusion and refinement methods, context-guided local inference methods, and shallow-deep network feature integration methods.

GLNet \cite{glnet} was the first to introduce the idea of global-local branch fusion, and achieved semantic segmentation on UHR images by designing a global-local-refinement scheme. GRNet \cite{grnet} adopted such a scheme and introduced a patch proposal subnetwork, enabling the network to adaptively refine regions of interest in UHR images. CascadePSP \cite{cascadepsp} achieved gradual refinement of segmentation masks by cascading multiple PSPNets \cite{pspnet}. PointRend \cite{pointrend} treated image segmentation as a rendering process and similarly achieved iterative refinement of segmentation masks. UHRSNet \cite{uhrsnet} improved the global-local feature fusion method and reduced the computational redundancy in GLNet \cite{glnet}. Similarly, SGHRQ \cite{sghrq} adopted a dual-branch architecture consisting of a semantic branch and a spatial branch, where the information extracted by the semantic branch is used as memory for the spatial branch to obtain high-resolution feature queries.

ISDNet \cite{isdnet} first proposed the two-branch design of shallow and deep networks, achieving effective integration of shallow and deep features. WSDNet \cite{urur} introduced wavelet transform into the deep branch, further improving the accuracy and efficiency of wide-area remote sensing image segmentation. On this basis, GPWFormer \cite{gpwformer} further adopted Wave-ViT \cite{wavevit} as its shallow branch and achieved better fusion of dual-stream features. RUE \cite{rue} introduced the resolution-biased uncertainty estimation method. BPT \cite{bpt} used only a single branch and directly takes the whole image as input. After convolutional processing, it introduced a Transformer-based fusion module and additionally derived a boundary segmentation branch to improve the boundary prediction capability of the model.

In addition, some methods focus on improving the basic cropping-and-sliding-window inference paradigm. FCtL \cite{fctl,fctlplus} leveraged short-range contextual information for local segmentation networks, achieving high segmentation accuracy. To bridge the large spatial gap between different scales in global-local branch methods, MagNet \cite{magnet} further introduced intermediate continuous scales between the global and local branches, forming a multi-scale cascaded refinement framework. WiCoNet \cite{wiconet} employed a local branch and a wide-context branch, and uses contextual attention to model the semantic dependency between the two branches, thereby enhancing the model's ability to exploit local contextual information.

Although global-local branch fusion and refinement methods and shallow-deep network feature integration methods have achieved strong performance on high-spatial-resolution remote sensing images with limited geographical coverage (e.g., DeepGlobe \cite{deepglobe}, Inria Aerial \cite{inriaaerial}, and URUR \cite{urur}), they remain less effective when applied to scene-complex UWA images. Existing context-guided local inference methods, while capable of processing UWA images, still exhibit limitations in long-range semantic dependency modeling. In this paper, we propose a novel UWA segmentation network based on scale-frustum representations, which can simultaneously handle refined local details and largely contiguous ground objects.

\subsection{Whole-Scene Remote Sensing Image Segmentation}

REST \cite{rest} was recently proposed, mainly targeting holistic semantic segmentation of whole-scene remote sensing imagery. It built a spatial parallel interaction mechanism across multiple GPUs to capture and leverage global contextual information. Although whole-scene segmentation is closely related to UWA segmentation, the two task settings emphasize different aspects. REST mainly focuses on enabling end-to-end full-image feature interaction, whereas this paper focuses on UWA scenes characterized by extremely wide geographical coverage, significantly varying object scales, and long-range semantic continuity. In addition, REST relies on distributed fusion of full-image features across multiple GPUs, which incurs substantial memory consumption and depends heavily on multi-GPU parallel computation. In contrast, our method aims to model local, short-range, and long-range observations around each target region under a single-GPU setting.

\begin{figure*}[t]
  \centering
   \includegraphics[width=1\linewidth]{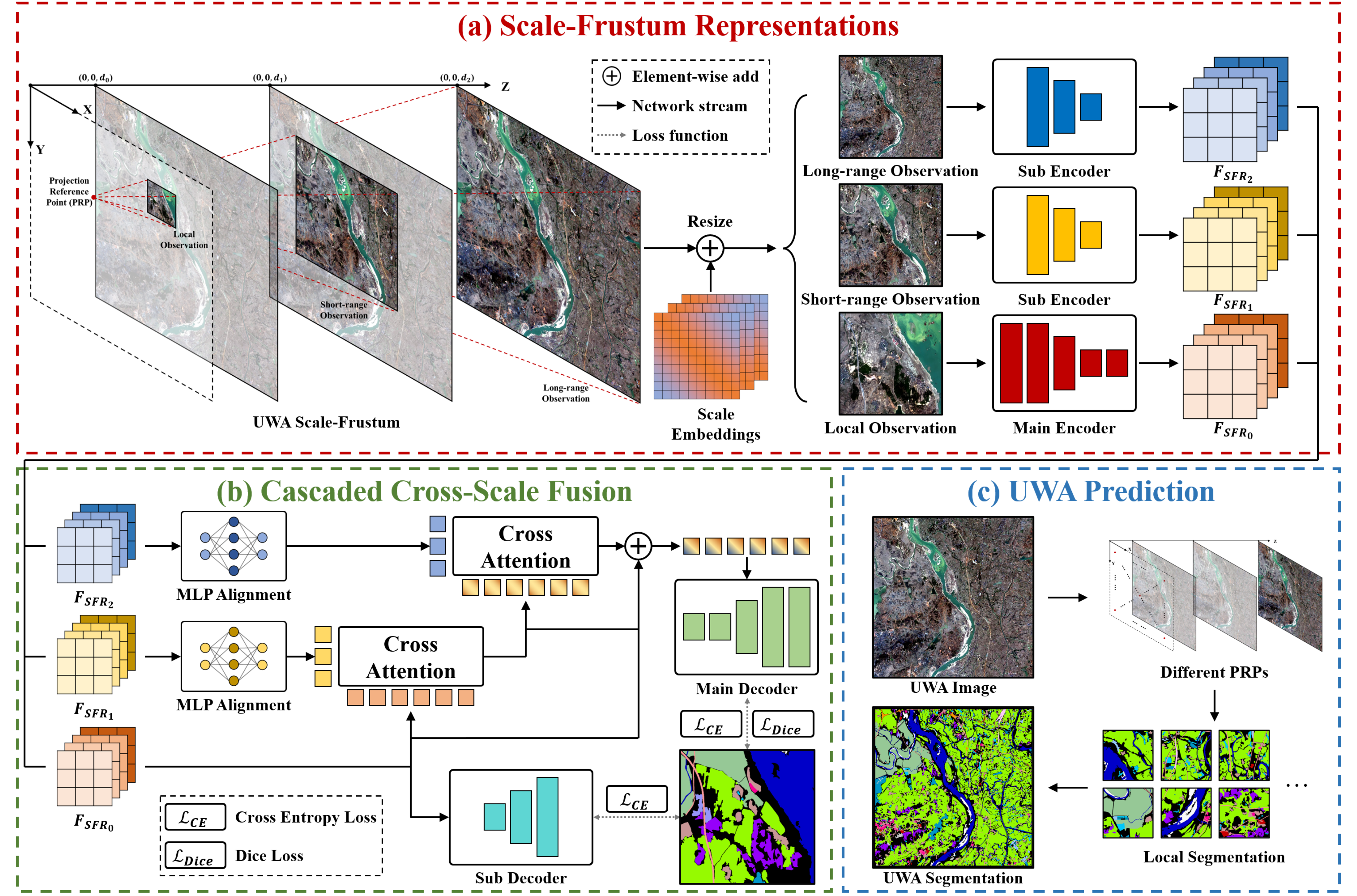}

   \caption{The proposed SFR-Net pipeline, designed to solve the dual UWA challenges. (a) First, Scale-Frustum Representations (SFR) are built to capture features across varying scales (local, short-range, and long-range). (b) Next, the Cascaded Cross-Scale Fusion (CCSF) module integrates these representations to maintain long-range contextual continuity. (c) Finally, the full UWA Prediction is generated by iteratively scanning the image with multiple Projection Reference Points (PRPs).}

   \label{fig:sfr-net}
\end{figure*}

%% file: sec/methodology.tex
\section{Methodology}

This section introduces the main components of the proposed framework. We first provide an overview of SFR-Net, and then present the construction of scale-frustum representations and the proposed cascaded cross-scale fusion module. Finally, the training and inference details as well as the implementation settings are described.

\subsection{Overview}

Ultra-wide area (UWA) remote sensing image segmentation requires the model to simultaneously recognize refined local objects and maintain long-range semantic continuity over extremely large geographical coverage. To address this issue, an intuitive solution is to introduce larger contextual observations for each local region. However, directly modeling the full-image context is computationally prohibitive, while simple patch-based processing inevitably fragments the semantic dependencies across distant regions. Therefore, the key question is how to construct a representation that can preserve refined local details while progressively incorporating short-range and long-range contextual information for the same target region.

Based on this motivation, we propose the Scale-Frustum Representation Network (SFR-Net). Inspired by the viewing frustums of remote sensing images captured from different altitudes, the core idea is to model the multi-scale contextual observations centered on the same local region as a scale-frustum representation. Specifically, given a projection reference point (PRP), we construct a set of observation windows with different spatial extents, corresponding to local, short-range, and long-range views. These windows provide complementary semantic cues at different scales. By organizing these observations in a unified frustum-style manner, the proposed representation naturally aligns multi-scale context around the same target region.

Based on the constructed scale-frustum representation, we further design a Cascaded Cross-Scale Fusion (CCSF) module to progressively inject contextual information from different scales into the local feature representation. Instead of directly merging all scales at once, CCSF performs cross-scale interaction in a cascaded manner, enabling the network to first absorb nearby contextual cues and then incorporate broader scene-level semantics. In this way, the final representation can preserve local discriminative details while benefiting from increasingly wider contextual support.

The overall pipeline of the proposed SFR-Net is illustrated in Fig.~\ref{fig:sfr-net}. First, for a given PRP, we construct the scale-frustum observations and resize them into a unified input size. These observations are then fed into a main encoder and several lightweight sub-encoders to extract multi-scale features. Next, the extracted features are aggregated by the proposed CCSF module, and the fused representation is passed to the main decoder for segmentation prediction, while an auxiliary sub-decoder is used to directly supervise the local branch. During inference, the whole UWA image is scanned with multiple PRPs, and the predictions of all local regions are merged to obtain the final full-image segmentation map.

\subsection{Scale-Frustum Representations}

The enormous geographical coverage of UWA images inherently encompasses continuous scales, from refined local details to global macroscopic overviews. This unique characteristic is similar to the viewing frustum of sensors at different altitudes during the imaging process. Different from conventional multi-scale cropping, the observations constructed in our framework are all centered on the same target region and preserve an explicit spatial correspondence across scales. In this paper, we model this as follows: for a specified PRP, we change its distance to the original image to obtain observation windows of different ranges and resize them to the same size, thereby constructing the UWA scale-frustum. In this way, the local region to be segmented and its surrounding contextual information at different ranges can be organized in a unified representation.

To better describe the proposed UWA scale-frustum, we first define a 3D coordinate system where the $X$-axis and $Y$-axis correspond to the spatial dimensions of the remote sensing image, and the $Z$-axis denotes the observation distance. Here, the introduced distance does not represent the actual sensor altitude, but serves as an abstract variable to characterize contextual observations with different spatial extents. Given a 2D remote sensing image $I\in \mathbb{R}^{H\times W\times C}$, where $H$, $W$, and $C$ represent height, width, and channel count, respectively, we extend it into 3D space by duplicating $I$ across $n$ designated distances $d_i (i=0,1,\ldots,n-1)$. This yields a set of 3D image slices ${I_i}$:

\begin{equation}
I_i(x, y, d_i) = I(x, y), \forall x \in [1, W], y \in [1, H]
\label{eq:uwa_replic}
\end{equation}

After that, we specify a projection reference point (PRP) $(w, h, 0)$, where $w\in [1, W]$ and $h\in [1, H]$. The PRP indicates the local region currently under analysis and serves as the common reference point for constructing multi-scale contextual observations. Consider an observation ray originating from the PRP and passing through a corner point $(x_c, y_c, d_{n-1})$ of the image on the $Z = d_{n-1}$ plane (e.g., the bottom-right corner $(W,H,d_{n-1})$). The parametric equation of this ray is:

\begin{equation}
\begin{cases}
x(t) = w + t \cdot (x_c - w) \\
y(t) = h + t \cdot (y_c - h) \\
z(t) = t \cdot d_{n-1}
\end{cases}, t \in [0,1]
\label{eq:ray}
\end{equation}

For a given plane $Z = d_i$, let $t_i = \frac{d_i}{d_{n-1}}$. The intersection point of the ray with this plane is:

\begin{equation}
(x_i , y_i , d_i) = (w + t_i \cdot (x_c - w), h + t_i \cdot (y_c - h), d_i)
\label{eq:inter_point}
\end{equation}

By defining such rays for all four corners of the $Z=d_{n-1}$ plane, we obtain a frustum-shaped observation window bounded by these intersection points. Intuitively, when the observation distance increases, the corresponding window covers a broader spatial range around the same PRP and thus provides richer contextual information. In contrast, smaller distances focus more on the refined local region and preserve more detailed spatial structures. Therefore, the resulting scale-frustum naturally establishes a unified description of local, short-range, and long-range observations for the same target region.

Subsequently, to enable the network to better perceive contextual features from different scales, we resize these observation windows to the same size (denoted as $\Omega_i$) and assign distinct learnable scale embeddings to different windows. Resizing all windows to a unified size makes it possible to process observations of different ranges within a consistent feature extraction framework, while the scale embeddings explicitly indicate their scale identities after resizing. Thereafter, similar to the approach in \cite{glnet, grnet, sghrq, rue}, we use a deep main encoder to process $\Omega_0$ and several shallow sub-encoders to process $\Omega_i \ (i=1,2,\ldots,n-1)$, thereby extracting contextual features of different scales, denoted as $F_{SFR_i}$. 

\begin{equation}
\begin{cases}
F_{SFR_0} = \mathrm{Main Encoder}(\Omega_0) \\
F_{SFR_i} = \mathrm{Sub Encoder}(\Omega_i) \quad , i \in {1, 2, \dots, n-1}
\end{cases}
\label{eq:cor_ray}
\end{equation}

Specifically, the local observation is processed by the main encoder to preserve refined local details, whereas the broader contextual observations are processed by lightweight sub-encoders to provide supplementary semantic cues with acceptable computational cost.

\subsection{Cascaded Cross-Scale Feature Fusion}

\begin{figure}[t]
  \centering
   \includegraphics[width=1\linewidth]{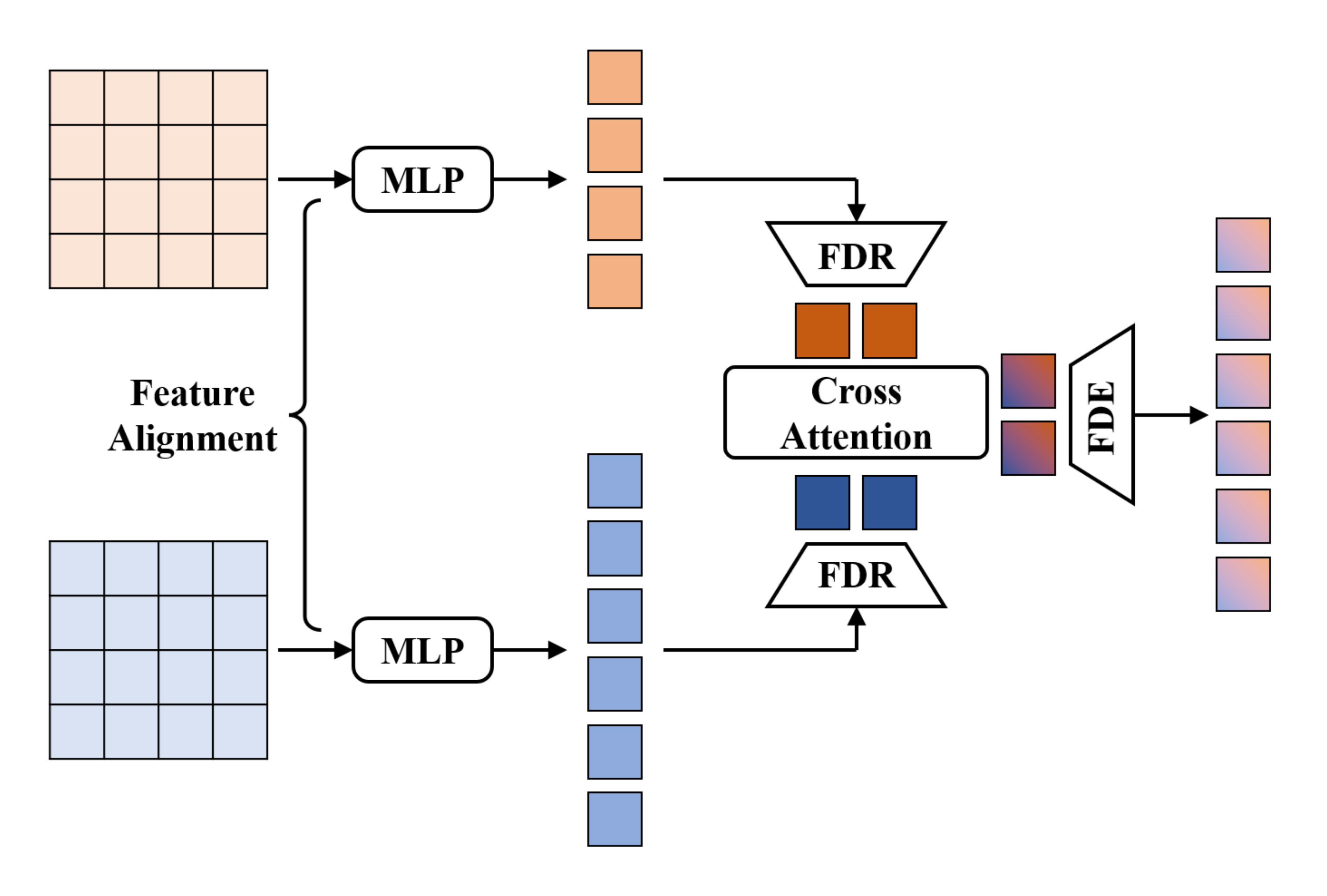}

   \caption{Illustration of a single fusion unit in the proposed cascaded cross-scale fusion module. Features from two adjacent scales are first aligned by MLP layers, then projected into a low-dimensional space through feature dimensionality reduction (FDR) for cross-attention computation, and finally restored by feature dimensionality expansion (FDE) for subsequent fusion.}
   \label{fig:fi module}
\end{figure}

Inspired by \cite{sghrq,vit}, we leverage the cross-attention mechanism to fuse features of different scales. Although the constructed scale-frustum representations provide complementary semantic cues at different ranges, their corresponding features differ in receptive field and semantic emphasis, making direct fusion suboptimal. In particular, the local observation preserves refined local details, while the broader observations provide increasingly wider contextual support. Therefore, an appropriate fusion mechanism should effectively introduce contextual semantics while preserving the discriminative capability of the local representation. To this end, we design a Cascaded Cross-Scale Fusion (CCSF) module to progressively inject multi-scale contextual information into the local feature representation. As illustrated in Fig.~\ref{fig:fi module}, a single fusion unit in CCSF consists of feature alignment, feature dimensionality reduction, cross-attention interaction, and feature dimensionality expansion.

For the obtained scale-frustum representations $F_{SFR_i}$ from different scales, we first use a multilayer perceptron (MLP) to achieve feature alignment with the local observation window. This step projects features from different scales into a more compatible representation space before cross-scale interaction. Before performing cross-attention computation, we further use a feature dimensionality reduction (FDR) module to obtain low-dimensional features (denoted as $F_{FDR_i}$), thereby reducing the computational burden of cross-attention while preserving the major semantic information. This process can be expressed as follows:

\begin{equation}
\begin{cases}
F_{FDR_0} = \mathrm{FDR}_0(F_{SFR_0}) \\
F_{FDR_i} = \mathrm{FDR}_i(\mathrm{MLP}(F_{SFR_i})) \quad , i \in {1, 2, \dots, n-1}
\end{cases}
\label{eq:mlp2fdr}
\end{equation}

Subsequently, cascaded cross-attention computation of different scales is performed. Instead of directly merging all scales at once, CCSF introduces contextual information in a progressive manner, allowing the network to first absorb relatively nearby contextual cues and then incorporate broader scene-level semantics. In this way, the fused representation can be gradually enhanced from local observation to larger contextual ranges. The resulting features (denoted as $F_{ca_i}$) can be expressed as follows:

\begin{equation}
F_{ca_i} = \mathrm{Softmax}\left(\frac{F_{FDR_{i-1}} \times F_{FDR_{i}}^{\top}}{\sqrt{dim}}\right) \times F_{FDR_{i}}
\label{eq:fdr2out}
\end{equation}

where ``$\times$'' denotes the matrix multiplication operation, $\mathrm{Softmax}$ represents the softmax activation function, and $dim$ stands for the reduced feature dimension. Here, the cross-attention mechanism enables features from a larger contextual range to interact with the feature from the previous scale, thereby progressively enriching the representation with complementary semantics.

The resulting features then pass through a feature dimensionality expansion (FDE) module to be restored to high-dimensional features. The final feature (denoted as $F_{fusion}$) is obtained by a weighted fusion of local features and the features computed via cross-attention:

\begin{equation}
F_{fusion} = F_{SFR_0} + \sum_{i=1}^{n-1} \alpha_i \cdot \mathrm{FDE}_i(F{ca_i})
\label{eq:finalfusion}
\end{equation}

This formulation preserves the local feature as the main representation while treating the contextual features from broader scales as complementary enhancements. As a result, the fused feature can simultaneously maintain refined local discriminability and incorporate broader contextual dependencies, which is particularly important for UWA segmentation.

\subsection{Training and Inference Strategy}

\begin{table*}[t]
\centering
\caption{Quantitative comparison on the GID and FBPS datasets. The best results are shown in bold and the second-best results are underlined.}
\label{tab:main_results}
\setlength{\tabcolsep}{7pt}
\renewcommand{\arraystretch}{1.15}
\begin{tabular}{l l cc cc c}
\hline
\multirow{2}{*}{Model} & \multirow{2}{*}{Backbone} & \multicolumn{2}{c}{GID} & \multicolumn{2}{c}{FBPS} & \multirow{2}{*}{MEM (MB)$\downarrow$} \\
 &  & mIoU (\%)$\uparrow$ & OA (\%)$\uparrow$ & mIoU (\%)$\uparrow$ & OA (\%)$\uparrow$ &  \\
\hline

\multicolumn{7}{l}{\footnotesize\textbf{Generic Semantic Segmentation}} \\
PSPNet \cite{pspnet} & ResNet101 \cite{resnet} & 63.03 & 79.55 & 64.48 & 89.02 & 1130 \\
DeepLabv3+ \cite{deeplabv3plus} & ResNet101 \cite{resnet} & 64.59 & 80.44 & 64.01 & 89.00 & 1098 \\
FCN \cite{fcn} & HRNet48 \cite{hrnet} & 64.78 & 81.54 & 64.84 & 88.95 & 974 \\
OCRNet \cite{ocrnet} & HRNet48 \cite{hrnet} & 67.00 & 82.70 & 63.91 & 89.33 & 1160 \\
PointRend \cite{pointrend} & ResNet101 \cite{resnet} & 64.16 & 80.19 & 60.93 & 88.16 & \underline{772} \\
STDC \cite{stdc} & STDC \cite{stdc} & 67.11 & 81.68 & 50.94 & 85.70 & \textbf{568} \\
SegFormer \cite{segformer} & MiT-B5 \cite{segformer} & 71.76 & 85.14 & 65.73 & 89.30 & 1000 \\
ConvNeXt \cite{convnext} & ConvNeXt \cite{convnext} & 72.08 & 85.45 & 59.72 & 88.13 & 2204 \\
Swin Transformer \cite{swin} & Swin-Large \cite{swin} & \underline{72.95} & \underline{85.85} & 69.68 & 90.51 & 2156 \\

\hline
\multicolumn{7}{l}{\footnotesize\textbf{Ultra-High Resolution Segmentation}} \\
GLNet \cite{glnet} & ResNet50 \cite{resnet} & - & - & 44.73 & - & - \\
FCtL+ \cite{fctlplus} & VGG16 \cite{vgg} & - & - & 48.28 & - & - \\
MagNet \cite{magnet} & ResNet50 \cite{resnet} & - & - & 44.20 & - & - \\
ISDNet \cite{isdnet} & ResNet18 \cite{resnet} & 54.14 & 74.26 & 47.78 & 85.62 & 7830 \\

\hline
\multicolumn{7}{l}{\footnotesize\textbf{Whole-Scene Remote Sensing Image Segmentation}} \\
REST \cite{rest} & Swin-Large \cite{swin} & - & - & \underline{72.95} & \underline{92.78} & 2114 \\

\hline
\multicolumn{7}{l}{\footnotesize\textbf{Ultra-Wide Area Segmentation}} \\
SFR-Net (ours) & Swin-Large \cite{swin} & \textbf{74.67} & \textbf{86.94} & \textbf{77.24} & \textbf{92.91} & 2314 \\

\hline
\end{tabular}
\end{table*}

During training, directly modeling the entire UWA image is computationally impractical. Therefore, instead of feeding the full image into the network, we randomly sample a projection reference point (PRP) from the image plane for each training iteration and construct its corresponding scale-frustum representation. In this way, each training sample corresponds to a local region together with its aligned short-range and long-range contextual observations. As training proceeds, different sampled PRPs gradually cover the entire UWA image, enabling the network to learn both refined local semantics and long-range contextual dependencies over the full scene.

During inference, we segment the whole UWA image by scanning it with multiple PRPs, as illustrated in the lower-right part of Fig.~\ref{fig:sfr-net}. Specifically, we first densely sample PRPs on the image plane in a sliding-window manner on the local observation plane. For each PRP, a scale-frustum representation is constructed and fed into the trained SFR-Net to obtain the prediction of the corresponding local region. By repeating this process over all PRPs, the network produces a set of local segmentation results that together cover the full UWA image. These local predictions are then merged to form the final full-image segmentation map.

To improve the continuity and stability of the final prediction, overlapping inference can be adopted during the scanning process. In this case, multiple local predictions may contribute to the same pixel location. For overlapping regions, we aggregate the logits from different predictions and assign the final category according to the maximum aggregated response. Such a strategy helps reduce boundary inconsistency between adjacent local predictions and further improves the semantic continuity of the final segmentation result.

For network supervision, we adopt a simple combination of dice loss and cross-entropy loss for the main decoder. The dice loss, denoted as $\mathcal{L}_{Dice}$, is used to improve the overlap quality between prediction and ground truth, while the cross-entropy loss for the main decoder, denoted as $\mathcal{L}_{CE_m}$, provides pixel-wise category supervision for the final segmentation output. In addition, we apply an auxiliary cross-entropy loss for the sub decoder, denoted as $\mathcal{L}_{CE_s}$, to directly supervise the local feature $F_{SFR_0}$. This auxiliary supervision encourages the local branch to preserve stronger local discriminative capability before cross-scale fusion. The total loss, denoted as $\mathcal{L}_{total}$, is defined as follows:

\begin{equation}
\mathcal{L}_{total} = \lambda_1 \mathcal{L}_{Dice} + \lambda_2 \mathcal{L}_{CE_m} + \lambda_3 \mathcal{L}_{CE_s},
\label{eq:total_loss}
\end{equation}

where $\lambda_1$, $\lambda_2$, and $\lambda_3$ denote the corresponding loss weights.

\subsection{Implementation Details}

Following prior work \cite{isdnet,rest}, we employ Swin-Large \cite{swin} as the main encoder, UperNet \cite{upernet} as the main decoder, ResNet-18 \cite{resnet} as the sub-encoder, and a simple FCN \cite{fcn} as the sub-decoder. In this design, the stronger main encoder-decoder pair is used to preserve refined local representation ability, while the lightweight sub-encoder and sub-decoder are adopted to extract and supervise broader contextual observations without introducing excessive additional complexity. All models are initialized with ImageNet pre-trained parameters. For all experiments, we set the batch size to 4 and train for a maximum of 320k iterations. To ensure a fair comparison and convergence, all competing methods are trained with these same settings. We use the AdamW \cite{adamw} optimizer with an initial learning rate of $6 \times 10^{-5}$ and a weight decay of 0.01. The learning rate is linearly increased during the first 1500 iterations as a warm-up phase.

For our SFR module, we set the distances to [1, 3, 14] by default. Specifically, the long-range distance is set to 14 because the ratio between the default local input size ($512 \times 512$) and the full UWA image size (approximately $7000 \times 7000$) is close to 1:14, allowing the long-range observation to cover nearly the full spatial extent of the input image. The short-range distance is set to 3 to provide an intermediate contextual range for capturing local surrounding semantics between the local and long-range observations. During inference, to compare with generic cropping-based methods, we use a uniform $512 \times 512$ sliding window with a stride of 128. Specifically, for ISDNet \cite{isdnet}, we follow its prescribed configuration: using a $2000 \times 2000$ crop size during training and processing $7000 \times 7000$ inputs directly at test time. For the loss function, we set $\lambda_1 = 5$, $\lambda_2 = 1$, and $\lambda_3 = 1$. This setting follows a common practice in semantic segmentation to assign a relatively larger weight to the dice term, which is also helpful for alleviating the impact of category imbalance in large-area remote sensing scenes. For data augmentation, we apply random horizontal and vertical flipping with a probability of 0.5, together with photometric distortion. The same augmentation strategy is applied to all competing methods for fair comparison.

Our experiments are implemented in PyTorch \cite{pytorch} using the MMsegmentation \cite{mmseg} toolbox. All experiments are conducted on a single NVIDIA RTX 4090 GPU. We report peak GPU memory (MEM) monitored via the \texttt{nvitop} command, and set the batch size to 1 for all methods during inference.

%% file: sec/experiments_and_analysis.tex
\section{Experiments and Analysis}

\subsection{Datasets and Evaluation Metrics}

We evaluate our method on two datasets that meet the UWA segmentation criteria: GID \cite{gid} and FBPS \cite{fbps}. We adopt widely used evaluation metrics for all experiments.

\textbf{GID.} The GID dataset contains 150 images, each with a size of $7300 \times 6900$ pixels, and includes 5 land cover categories. The Ground Sampling Distance (GSD) is 3.24~m, and each image covers 529~km$^2$. Following the standard split from \cite{gid}, we use 120 images for training and 30 images for validation and testing. As a representative UWA dataset with relatively coarse land-cover categories, GID is suitable for evaluating the overall segmentation performance of different methods under the UWA setting.

\textbf{FBPS.} The Five-Billion-Pixels (FBPS) dataset also contains 150 images and shares the same image set, spatial resolution, and geographical coverage as GID. The key difference is that FBPS provides finer-grained annotations with 24 land cover categories. Following \cite{fbps}, we use the same 120/30 split for training and validation/testing. Owing to its finer-grained annotation and more complex category composition, FBPS is particularly suitable for further evaluating the capability of different methods in handling category confusion and class imbalance in UWA segmentation.

\textbf{Evaluation Metrics.} We mainly report mean Intersection over Union (mIoU) and Overall Accuracy (OA) for comparisons, and additionally report mean F1-score (mF1) in ablation studies. In terms of computational cost, we mainly report peak GPU memory consumption (MEM). In addition, we also compare FLOPs, parameter count (Params), and inference speed (FPS) with representative baseline and state-of-the-art methods.

\subsection{Experiments on the GID Dataset}

\begin{figure*}[t]
  \centering
   \includegraphics[width=1\linewidth]{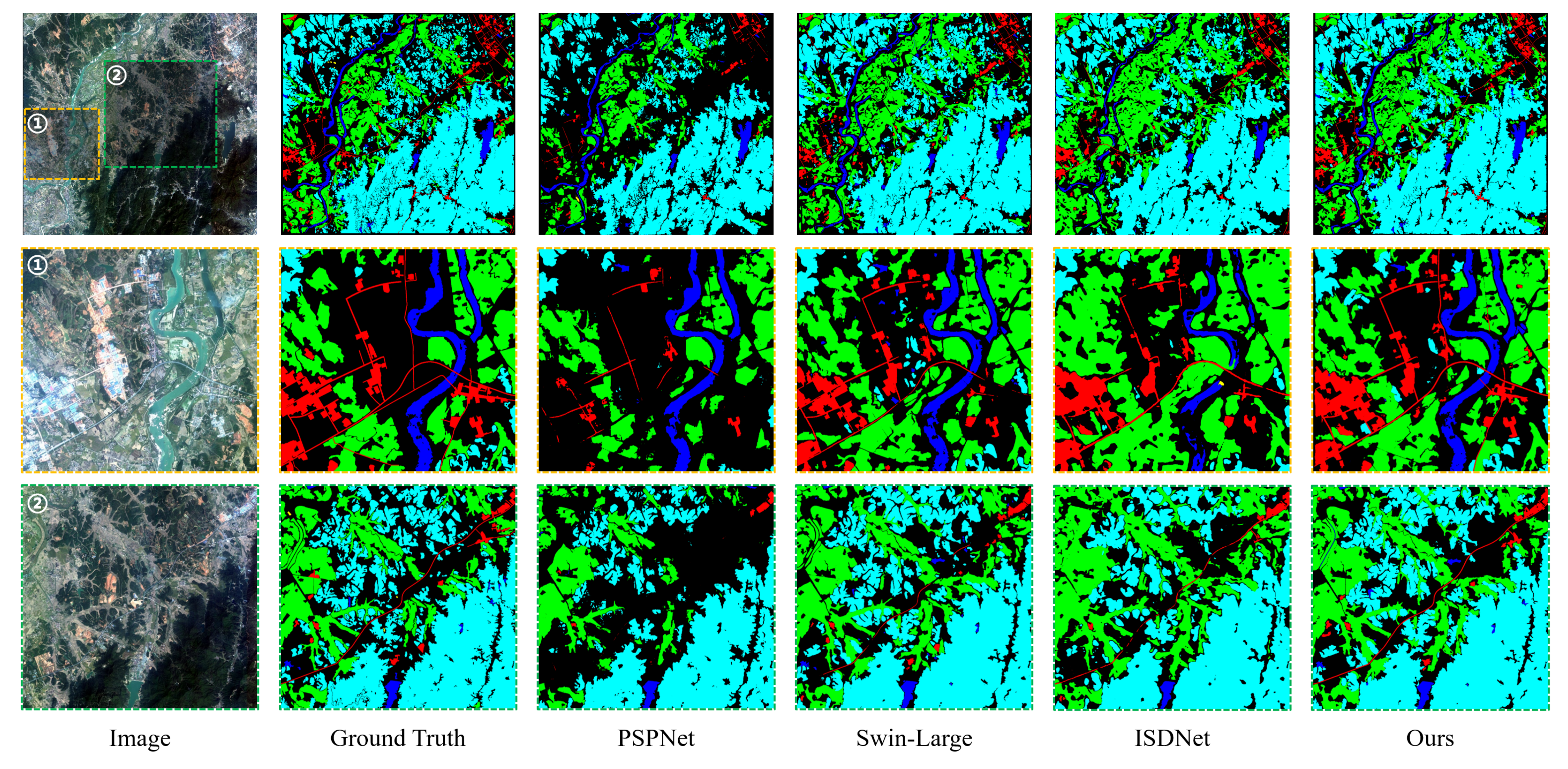}

   \caption{Qualitative comparison of segmentation results on the GID dataset. The first column displays the full UWA image and corresponding zoomed-in observation windows. Compared with existing methods, our method not only captures refined local details (e.g., small buildings) but also better maintains long-range contextual consistency (e.g., the continuity of roads).}
   \label{gid_comp}
\end{figure*}

We compared our method with several generic segmentation networks and ultra-high resolution segmentation methods on the GID dataset. Table~\ref{tab:main_results} presents the quantitative results in terms of accuracy and memory overhead. Overall, our method achieves the highest mIoU and OA without introducing excessive memory overhead. Specifically, our method outperforms the previous state-of-the-art method by 1.72\% in mIoU and 1.09\% in OA.

Fig.~\ref{gid_comp} shows the qualitative visualization results. It can be observed that, on the one hand, generic segmentation methods can segment refined local details within their observation window, but struggle to maintain semantic continuity in long-range observation. For example, in the 4th column of Fig.~\ref{gid_comp}, the Swin Transformer \cite{swin} exhibits discontinuous segments in a road spanning a long range. On the other hand, ultra-high resolution segmentation methods process the entire image directly, such as ISDNet \cite{isdnet} in the 5th column of Fig.~\ref{gid_comp}. Although these methods can better handle semantic information in long-range observation, they fail to capture refined local details, e.g., missing details of local houses and roads. In contrast, our method, by incorporating scale-frustum representations, simultaneously captures rich refined details and maintains better semantic continuity in the long-range observation.

\subsection{Experiments on the FBPS Dataset}

To verify the performance of our method on datasets characterized by fine-grained classes and severe class imbalance, we compared our method with existing methods on the FBPS dataset. Table~\ref{tab:main_results} presents the quantitative results in terms of accuracy and memory. In general, our method achieves significantly higher mIoU and OA. Compared with the previous state-of-the-art method, our method achieves gains of 4.29\% in mIoU and 0.13\% in OA.

\begin{figure*}[t]
  \centering
   \includegraphics[width=1\linewidth]{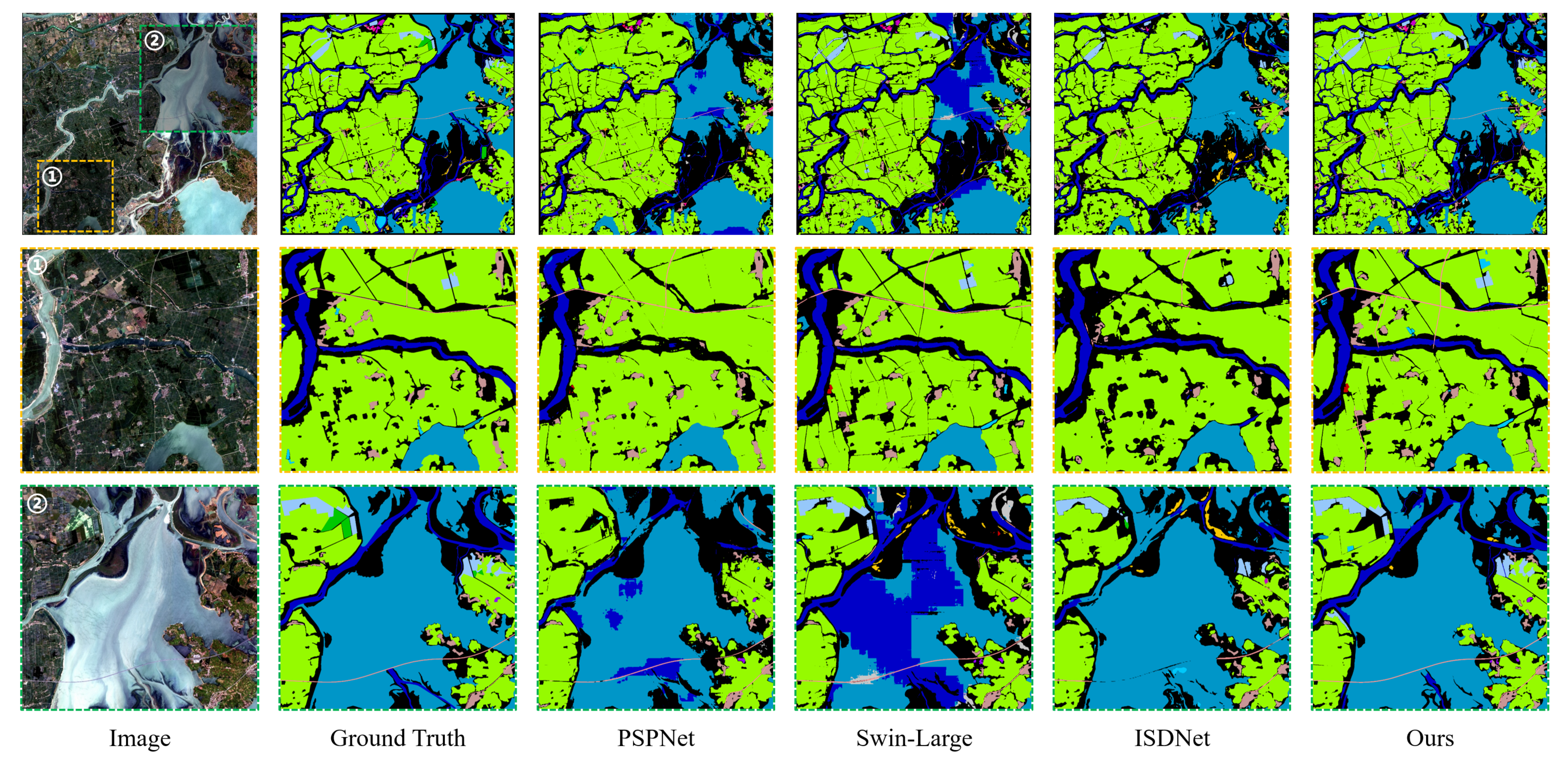}

   \caption{Qualitative comparison of segmentation results on the FBPS dataset. The first column displays the full UWA image and corresponding zoomed-in observation windows. Our method demonstrates superior performance in handling semantically confusable classes (e.g., distinguishing ``river'' from ``pond''), which appear similar in local patches but differ in their large-scale contextual structure.}
   \label{fbps_comp}
\end{figure*}

Fig.~\ref{fbps_comp} shows the visualization results. It can be observed that generic segmentation methods struggle to distinguish semantically confusable classes that are vastly different in scale but appear similar within a local observation window. For example, in the 4th column of Fig.~\ref{fbps_comp}, the Swin Transformer \cite{swin} finds it hard to differentiate between ``river'' and ``pond''. These two classes appear highly similar within a local observation window; however, the former exhibits a linear structure, while the latter has a large contiguous shape. This reliance on local-scale features makes it difficult for generic methods to distinguish them.

\begin{table}[t]
    \centering
    \caption{Comparison of computational cost and segmentation performance on the FBPS dataset. The best results are shown in bold.}
    \label{tab:efficiency}
    \setlength{\tabcolsep}{3pt}
    \renewcommand{\arraystretch}{1.1}
    \resizebox{\linewidth}{!}{
    \begin{tabular}{lccccc}
    \toprule
    Models & mIoU (\%)$\uparrow$ & FLOPs (G)$\downarrow$ & Params (M)$\downarrow$ & Mem (MB)$\downarrow$ & FPS$\uparrow$ \\
    \midrule
    Swin Transformer \cite{swin} & 69.68 & \textbf{588.1} & \textbf{233.85} & \textbf{2156} & \textbf{0.11} \\
    REST \cite{rest} & 72.95 & 627.0 & 272.00 & - & - \\
    SFR-Net (ours) & \textbf{77.24} & 839.8 & 271.11 & 2314 & 0.04 \\
    \bottomrule
    \end{tabular}
    }
\end{table}

For UHR segmentation methods, the increased number of classes makes it even harder to capture refined local details. Taking ISDNet \cite{isdnet} as an example, it tends to predict only the class with a higher overall probability of occurrence on confusable classes, further degrading segmentation accuracy. Moreover, such methods struggle to segment long-range ground objects, such as the road crossing the pond in the 5th column of Fig.~\ref{fbps_comp}. This further degrades segmentation performance, which significantly limits their performance on UWA segmentation.

In contrast, our method simultaneously balances refined local details and long-range semantic continuity. Additionally, the scale-frustum representations provide the multi-scale context necessary to distinguish these confusable classes, which is particularly important for the FBPS dataset with finer-grained categories and more severe class imbalance.

\subsection{Computational Cost Analysis}

To further analyze the computational cost of the proposed method, we compare SFR-Net with baseline methods on the FBPS dataset in terms of segmentation accuracy, FLOPs, parameter count, peak GPU memory consumption, and inference speed. The results are summarized in Table~\ref{tab:efficiency}.

As shown in Table~\ref{tab:efficiency}, SFR-Net achieves the best segmentation performance on FBPS, while this gain is accompanied by increased computational cost. Compared with REST, SFR-Net maintains a comparable parameter count with only a moderate increase in FLOPs, indicating that the performance improvement mainly comes from the proposed architecture rather than simply scaling up the model size.

\subsection{Ablation Study}

In this section, we conduct comprehensive ablation studies to validate the effectiveness of our proposed method and its components.

\subsubsection{Effectiveness of scale-frustum representations}

\begin{table}
  \caption{Ablation study on the effectiveness of the proposed Scale-Frustum Representations (SFR) on the GID dataset. ``Baseline'' denotes the segmentation model without SFR, while ``$\checkmark$'' indicates the same model augmented with our SFR.}
  \centering
  \begin{tabular}{c c c c c}
    \toprule
    Baseline & SFR & mIoU(\%)$\uparrow$ & OA(\%)$\uparrow$ & mF1(\%)$\uparrow$ \\
    \midrule
    \multirow{3}{*}{PSPNet \cite{pspnet}} 
      &  & 63.03 & 79.55 & 77.24 \\
      & \multirow{2}{*}{$\checkmark$} & 71.75 & 84.70 & 84.46 \\
      &  & \textbf{(+8.72)} & \textbf{(+5.15)} & \textbf{(+7.22)} \\
    \midrule
    \multirow{3}{*}{DeepLabv3+ \cite{deeplabv3plus}} 
      &  & 64.59 & 80.44 & 78.69 \\
      & \multirow{2}{*}{$\checkmark$} & 72.71 & 86.02 & 84.41 \\
      &  & \textbf{(+8.12)} & \textbf{(+5.58)} & \textbf{(+5.72)} \\
    \midrule
    \multirow{3}{*}{UperNet \cite{upernet}}  
      &  & 72.95 & 85.85 & 84.68 \\
      & \multirow{2}{*}{$\checkmark$} & 74.67 & 86.94 & 85.88 \\
      &  & \textbf{(+1.72)} & \textbf{(+1.09)} & \textbf{(+1.20)} \\
    \bottomrule
  \end{tabular}
  \label{tab:ab_sfr}
\end{table}

\begin{figure}[t]
  \centering
   \includegraphics[width=1\linewidth]{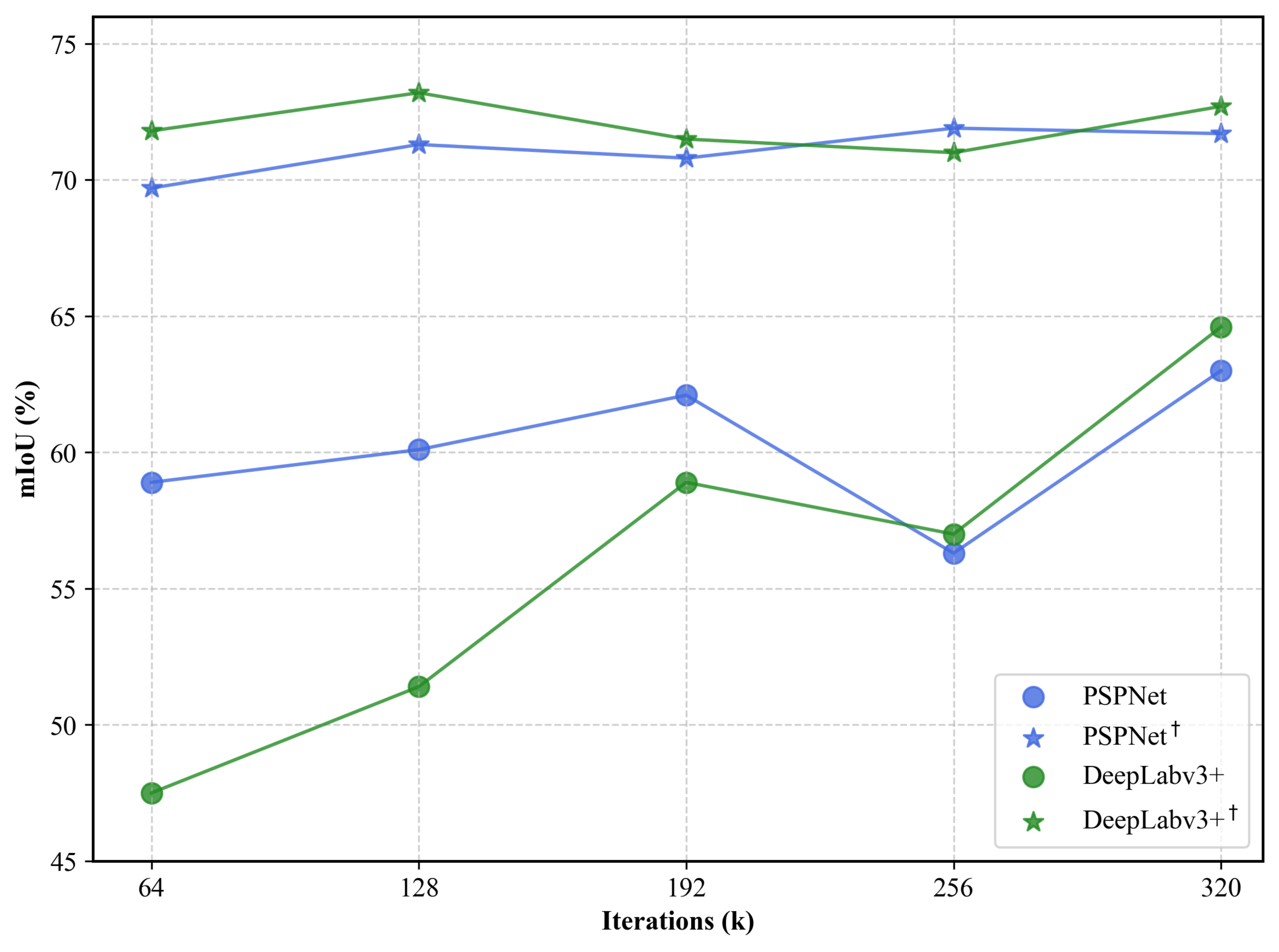}

   \caption{Impact of Scale-Frustum Representations (SFR) on accuracy and convergence speed on the GID test set. Models augmented with SFR (denoted by ``$\dagger$'') consistently achieve higher mIoU and converge significantly faster.}
   \label{fig:gid_by_iter}
\end{figure}

We validated the general-purpose utility of our scale-frustum representations (SFR) by applying them to several generic segmentation networks. The results are presented in Table~\ref{tab:ab_sfr}. These results show that SFR consistently improves UWA segmentation accuracy across all tested architectures.
Notably, the performance gains are most significant for models that originally struggle with UWA segmentation; for instance, applying SFR to PSPNet boosts its mIoU and OA by 8.72\% and 5.15\%, respectively. We also analyzed the impact on convergence speed. As shown in Fig.~\ref{fig:gid_by_iter}, SFR not only improves accuracy but also accelerates convergence, achieving up to +7.19\% mIoU with only 20\% of the total training iterations.
This demonstrates that the proposed SFR is a versatile module that can be seamlessly integrated with generic approaches, adapting them for the challenging UWA segmentation task.

\subsubsection{Effectiveness of cascaded cross-scale fusion}

\begin{table}
  \caption{Ablation study for the cascaded cross-scale fusion (CCSF) module on the GID dataset. The baseline is progressively enhanced with ``Short-range'' and ``Long-range'' Observation.}
  \centering
  \begin{tabular}{C{1.4cm} C{1.4cm} C{1.4cm} C{0.7cm} C{0.7cm} C{0.7cm}}
    \toprule
    Local & Short-range & Long-range & mIoU & OA & mF1 \\
    Observation & Observation & Observation & (\%)$\uparrow$ & (\%)$\uparrow$ & (\%)$\uparrow$ \\
    \midrule
    $\checkmark$ & & & 72.95 & 85.85 & 84.68 \\
    $\checkmark$ & $\checkmark$ & & 74.54 & 86.92 & 85.73 \\
    $\checkmark$ & & $\checkmark$ & 74.64 & 86.86 & \textbf{85.89} \\
    $\checkmark$ & $\checkmark$ & $\checkmark$ & \textbf{74.67} & \textbf{86.94} & 85.88 \\
    \bottomrule
  \end{tabular}
  
  \label{tab:ab_context_type}
\end{table}

\begin{table}
  \caption{Ablation study for the cascaded cross-scale fusion (CCSF) module on the FBPS dataset. The baseline is progressively enhanced with ``Short-range'' and ``Long-range'' Observation.}
  \centering
  \begin{tabular}{C{1.4cm} C{1.4cm} C{1.4cm} C{0.7cm} C{0.7cm} C{0.7cm}}
    \toprule
    Local & Short-range & Long-range & mIoU & OA & mF1 \\
    Observation & Observation & Observation & (\%)$\uparrow$ & (\%)$\uparrow$ & (\%)$\uparrow$ \\
    \midrule
    $\checkmark$ & & & 70.37 & 88.93 & 58.33 \\
    $\checkmark$ & $\checkmark$ & & 76.76 & 92.47 & 66.08 \\
    $\checkmark$ & & $\checkmark$ & 76.84 & 92.47 & \textbf{66.49} \\
    $\checkmark$ & $\checkmark$ & $\checkmark$ & \textbf{77.24} & \textbf{92.91} & 66.32 \\
    \bottomrule
  \end{tabular}
  
  \label{tab:ab_context_type_fbps}
\end{table}

We conducted ablation experiments on the Cascaded Cross-Scale Fusion (CCSF) module; the quantitative results are presented in Table~\ref{tab:ab_context_type} and Table~\ref{tab:ab_context_type_fbps}. As shown, the baseline model (using only the ``Local Observation'') yields the lowest performance. Introducing either the ``Short-range'' or ``Long-range'' observation window individually improves segmentation accuracy. The best mIoU and OA are achieved when all three scales—local, short-range, and long-range—are fused, while the mF1 remains comparable to the best value. These experiments demonstrate that for effective UWA segmentation, both the local semantic correlations (from short-range observation) and the global contextual awareness (from long-range observation) are indispensable.

\begin{figure}[t]
  \centering
   \includegraphics[width=1\linewidth]{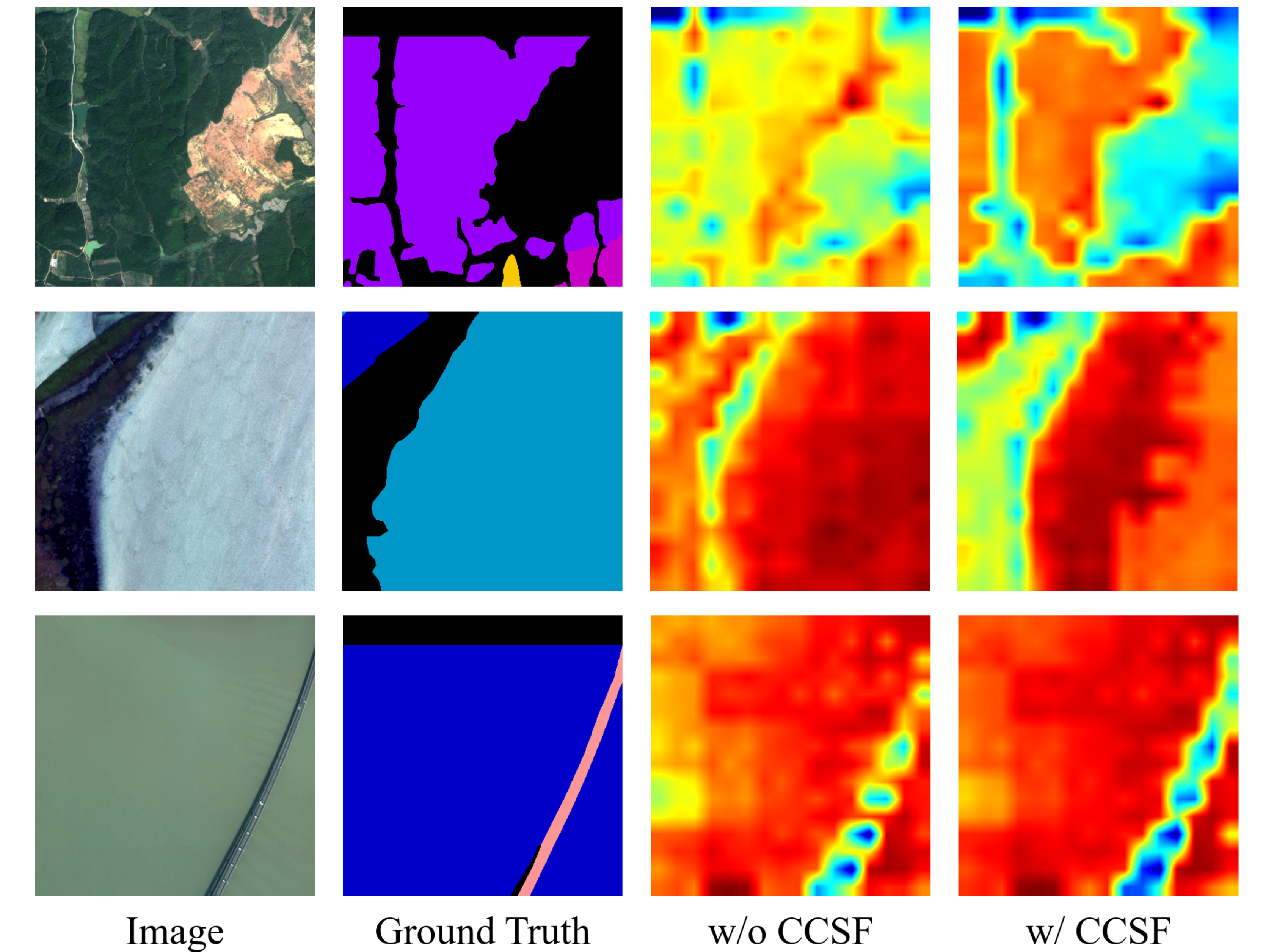}

   \caption{Qualitative comparison demonstrating the effectiveness of the CCSF module. The feature maps with CCSF show visibly sharper and more precise edges, enhancing local semantic continuity and foreground-background separation compared to the baseline without CCSF.}
   \label{fig:cfi_comp}
\end{figure}

Furthermore, Fig.~\ref{fig:cfi_comp} provides a qualitative visualization of the encoder's feature maps, comparing the baseline (without CCSF) to our full model (with CCSF). The CCSF module clearly enables the model to extract sharper and more precise edge features. In addition, the incorporated context helps to better distinguish background regions from foreground objects and improves local semantic continuity.

\subsubsection{Analysis of overlapping inference}

\begin{table}

  \caption{Analysis of inference with and without overlapping patches. The $\Delta$ column shows the mIoU gain from overlap, where a smaller value indicates better internal semantic continuity. Our SFR-Net's minimal $\Delta$ demonstrates superior robustness.}
  \centering
  \begin{tabular}{c c c c}
    \toprule
    \multirow{2}{*}{Models} & mIoU w/o & mIoU w/ & $\Delta$ \\
    & overlap (\%)$\uparrow$ & overlap (\%)$\uparrow$ & (\%)$\downarrow$ \\
    \midrule
    PSPNet \cite{pspnet}   & 61.77 & 63.03   & 1.26 \\
    DeepLabv3+ \cite{deeplabv3plus} & 63.01  & 64.59 & 1.58 \\
    FCN \cite{fcn}      & 63.03   & 64.78 & 1.75 \\
    STDC \cite{stdc}     & 65.84 & 67.11 & 1.27 \\
    SegFormer \cite{segformer} & 70.10 & 71.76 & 1.66 \\
    ConvNeXt \cite{convnext} & 70.85 & 72.08 & 1.23 \\
    Swin-L \cite{swin}     & 71.08 & 72.95 & 1.87 \\
    \textbf{SFR-Net(ours)}    & \textbf{73.83} & \textbf{74.67} & \textbf{0.84} \\
    \bottomrule
  \end{tabular}
  
  \label{tab:ab_stride}
\end{table}

We adopted two inference strategies: one using non-overlapping $512 \times 512$ tiles (a stride of 512) and another using overlapping tiles (a stride of 128). For overlapping regions, we aggregated predictions by summing the segmentation logits and selecting the class with the maximum value. We then compared the impact of these strategies on different models, with the results presented in Table~\ref{tab:ab_stride}. While all methods benefit from overlapping inference, our SFR-Net exhibits the smallest performance gap ($\Delta$) between the two strategies. This indicates that the proposed scale-frustum representations already provide superior semantic continuity across local patches, thereby reducing the reliance on the auxiliary improvements gained from overlapping sliding windows.

\subsubsection{Effect of the Number of Distances}

We further analyze the effect of the number of distances used in the proposed scale-frustum representation. We compare the default three-distance setting with a two-distance variant and variants containing more intermediate distances while keeping all other settings unchanged. The results are reported in Table~\ref{tab:distance_number_ablation}.

As shown in Table~\ref{tab:distance_number_ablation}, the default three-distance setting [1, 3, 14] achieves the best overall performance. When more distances are introduced, the segmentation accuracy does not improve further and even drops slightly in terms of mIoU and mF1. This indicates that the proposed local-short-long triplet is already sufficient to capture the required multi-scale contextual information for UWA segmentation. In contrast, adding more intermediate distances mainly introduces redundant observations and additional computational burden, without providing more effective semantic guidance.

\begin{table}[t]
\centering
\caption{Ablation study on different numbers of distances.}
\label{tab:distance_number_ablation}
\setlength{\tabcolsep}{7pt}
\renewcommand{\arraystretch}{1.1}
\begin{tabular}{cccc}
\toprule
Distances & mIoU (\%)$\uparrow$ & OA (\%)$\uparrow$ & mF1 (\%)$\uparrow$ \\
\midrule
$[1, 3]$ & 74.54 & 86.92 & 85.73 \\
$[1, 3, 14]$ & \textbf{74.67} & \textbf{86.94} & \textbf{85.88} \\
$[1, 3, 6, 14]$ & 74.41 & 86.78 & 85.71 \\
$[1, 3, 6, 10, 14]$ & 74.38 & 86.87 & 85.59 \\
\bottomrule
\end{tabular}
\end{table}

\subsubsection{Effect of the Intermediate Distance Setting}

We further analyze the effect of the intermediate distance setting in the proposed scale-frustum representation. Specifically, we fix the local observation distance and the long-range observation distance as 1 and 14, respectively, and vary only the short-range distance. The results are reported in Table~\ref{tab:distance_ablation}.

As shown in Table~\ref{tab:distance_ablation}, the configuration of [1, 3, 14] achieves the best overall performance. When the short-range distance deviates from this setting, the segmentation accuracy drops slightly. This phenomenon can be explained from two aspects. On the one hand, an excessively small short-range distance leads to a field of view that highly overlaps with the local observation, resulting in redundant information and insufficient distinct contextual guidance. On the other hand, an excessively large short-range distance weakens the specific contextual cues required for refining local boundaries, making it more difficult for the network to effectively adjust local predictions. These results indicate that the proposed intermediate distance setting provides a suitable balance between local refinement and contextual perception.

\begin{table}[t]
\centering
\caption{Ablation study on different intermediate distance settings.}
\label{tab:distance_ablation}
\setlength{\tabcolsep}{8pt}
\renewcommand{\arraystretch}{1.1}
\begin{tabular}{cccc}
\toprule
Distances & mIoU (\%)$\uparrow$ & OA (\%)$\uparrow$ & mF1 (\%)$\uparrow$ \\
\midrule
$[1, 2, 14]$ & 74.62 & 86.87 & 85.86\\
$[1, 3, 14]$ & \textbf{74.67} & \textbf{86.94} & \textbf{85.88}\\
$[1, 4, 14]$ & 74.35 & 86.78 & 85.63\\
$[1, 5, 14]$ & 74.15 & 86.72 & 85.52\\
\bottomrule
\end{tabular}
\end{table}

\subsubsection{Analysis of combined loss}

We further analyze the effect of the adopted loss formulation. Specifically, we compare the full loss setting with a variant using only cross-entropy supervision, while keeping all other settings unchanged. The results are reported in Table~\ref{tab:loss_ablation}.

As shown in Table~\ref{tab:loss_ablation}, using only cross-entropy loss already achieves competitive performance, indicating that the proposed framework itself is effective. After further introducing the dice loss, the segmentation accuracy is consistently improved in terms of mIoU, OA, and mF1. This suggests that the dice term provides beneficial complementary supervision, which is helpful for optimizing segmentation quality in UWA scenes.

\begin{figure*}[t]
\centering
\includegraphics[width=\linewidth]{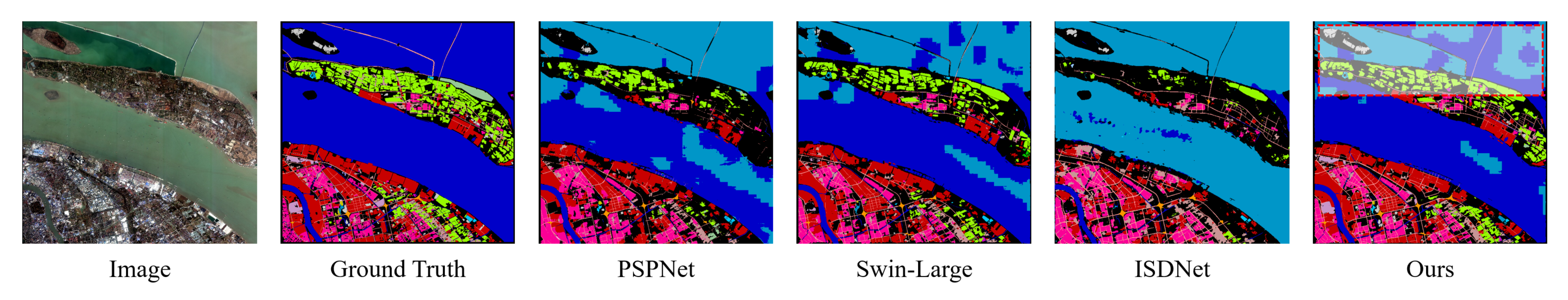}
\caption{Visualization of failure cases. Most methods fail to perfectly distinguish semantically similar water bodies, such as ``river'' and ``lake''. Although scale-frustum representations provide broader contextual information, our method may still struggle when visual appearances and annotation boundaries are ambiguous.}
\label{fig:failure}
\end{figure*}

\begin{table}[t]
\centering
\caption{Ablation study on the effect of dice loss.}
\label{tab:loss_ablation}
\setlength{\tabcolsep}{6pt}
\renewcommand{\arraystretch}{1.1}
\begin{tabular}{ccccc}
\toprule
CE Loss & Dice Loss & mIoU (\%)$\uparrow$ & OA (\%)$\uparrow$ & mF1 (\%)$\uparrow$ \\
\midrule
\checkmark &  & 74.45 & 86.77 & 85.75 \\
\checkmark & \checkmark & \textbf{74.67} & \textbf{86.94} & \textbf{85.88} \\
\bottomrule
\end{tabular}
\end{table}

\subsection{Discussion and Limitation}

Although the proposed SFR-Net achieves state-of-the-art performance on both the GID and FBPS datasets, it still has some limitations. As shown in Fig.~\ref{fig:failure}, semantically similar water bodies, such as ``river'' and ``lake'', remain challenging for existing methods. Most compared methods tend to rely on local appearance and produce fragmented or inconsistent predictions. By incorporating scale-frustum representations, SFR-Net can exploit broader contextual cues and generally provides more coherent predictions. Nevertheless, it may still fail when the visual appearance is highly similar and the semantic boundary is ambiguous.

This phenomenon indicates that UWA segmentation requires not only larger contextual observations, but also more reliable semantic reasoning for ambiguous land-cover categories. Such ambiguity may come from two aspects. First, some categories are inherently difficult to distinguish from local appearance alone and may depend on global topology or hydrological structure. Second, inconsistent annotation standards in large-scale datasets may introduce uncertain semantic boundaries. In addition, SFR-Net introduces additional computational cost due to multi-scale observation construction and cross-scale fusion. Future work will therefore focus on robust learning under ambiguous annotations, more adaptive scale selection, and more efficient cross-scale interaction.

%% file: sec/conclusion.tex
\section{Conclusion}

In this paper, we introduced the ultra-wide area (UWA) remote sensing image segmentation task, which requires the model to simultaneously handle ground objects with significantly varying scales and maintain long-range contextual semantic continuity over extremely large geographical coverage. To address these challenges, we proposed the Scale-Frustum Representation Network (SFR-Net), which constructs scale-frustum representations for aligned multi-scale contextual modeling and introduces a cascaded cross-scale fusion module for progressive context integration.

Extensive experiments on the GID and FBPS datasets demonstrate the effectiveness of the proposed method. SFR-Net achieves the best performance among the compared methods, improving mIoU by 1.72\% on GID and 4.29\% on FBPS over the strongest competing methods. The ablation studies further validate the effectiveness of the proposed scale-frustum representations, cascaded cross-scale fusion, distance settings, overlapping inference strategy, and combined loss formulation. In addition, the results show that SFR can be integrated with generic segmentation networks to improve their UWA segmentation performance and convergence speed.

In the future, we will further explore more adaptive and efficient UWA segmentation methods, including dynamic scale selection, lightweight cross-scale interaction, and more robust learning strategies for ambiguous land-cover categories. We expect this work to motivate further research on scalable contextual modeling for ultra-wide area remote sensing image analysis.


%





\ifCLASSOPTIONcaptionsoff
  \newpage
\fi




%% file: sec/biography.tex
\bibliographystyle{IEEEtran}
\bibliography{references.bib}

\begin{IEEEbiography}[{\includegraphics[width=1in,height=1.25in,clip,keepaspectratio]{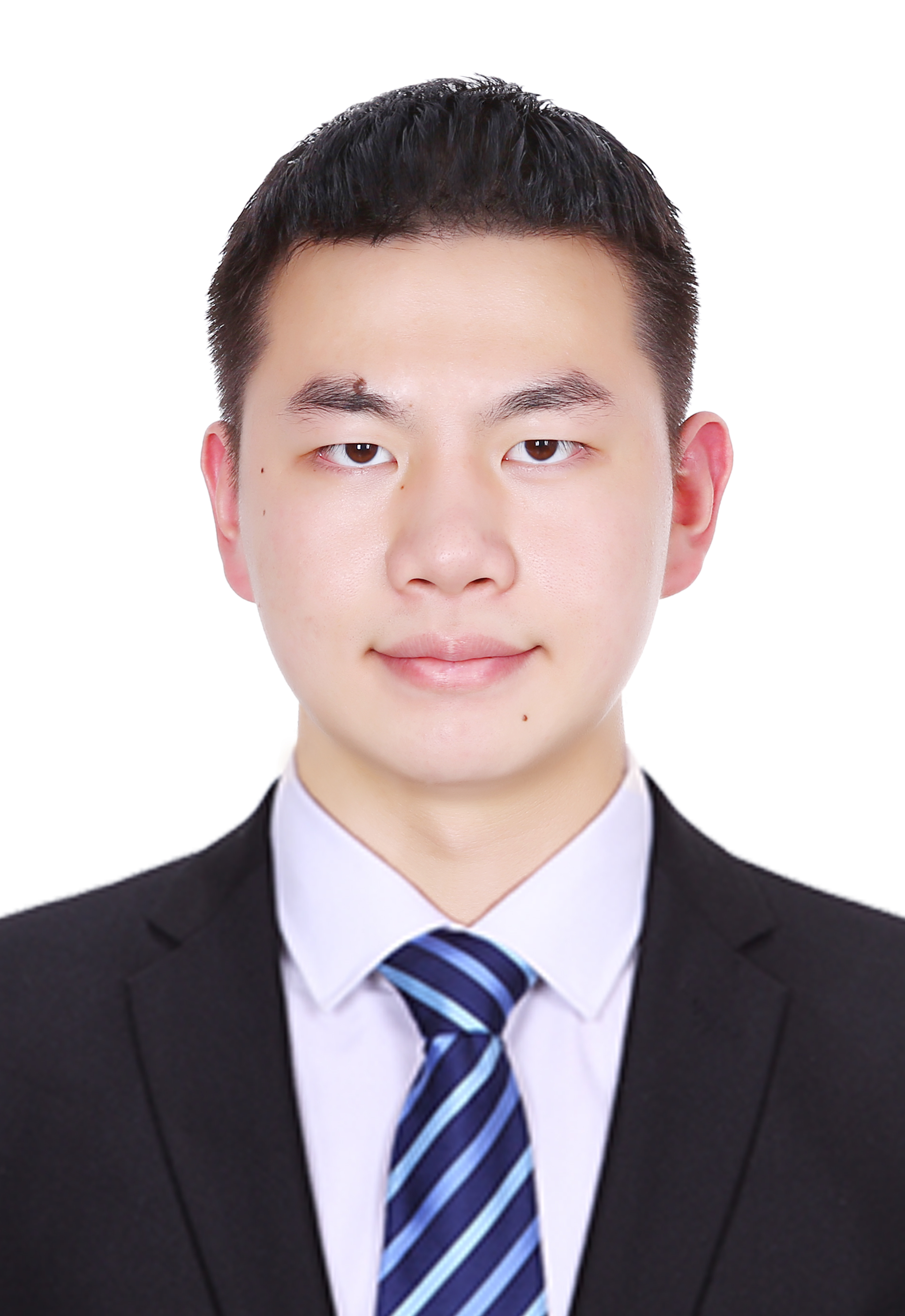}}]
{Chuyu Zhong}

received the B.S. degree from the School of Astronautics, Beihang University, Beijing, China, in 2025. He is currently pursuing the Ph.D. degree with the Image Processing Center, School of Astronautics, Beihang University. 

His research interests include image processing and deep learning, particularly ultra-wide-area and ultra-high-resolution remote sensing image processing.

\end{IEEEbiography}

\begin{IEEEbiography}[{\includegraphics[width=1in,height=1.25in,clip,keepaspectratio]{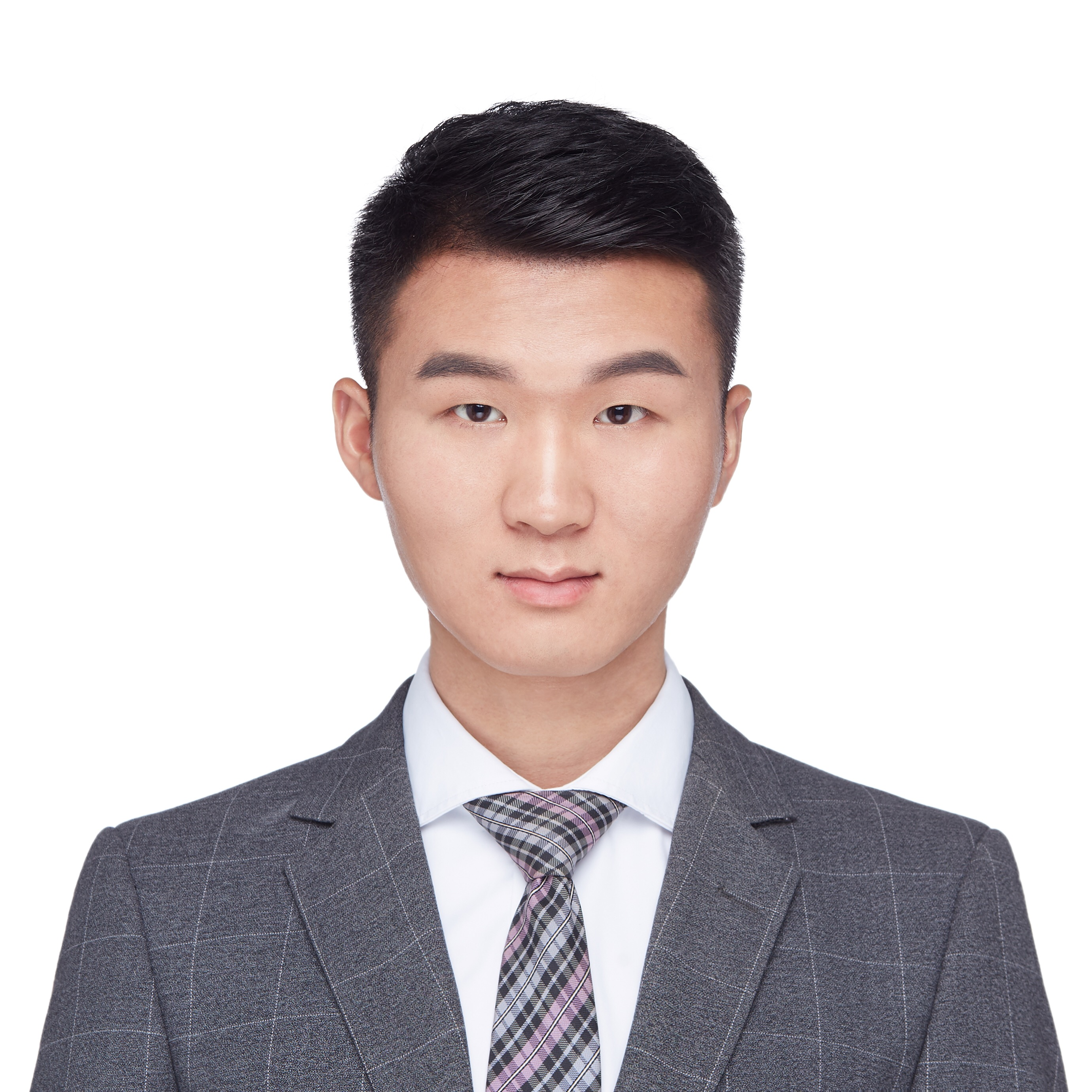}}]{Keyan Chen \text{(Member, IEEE)}} 
is a Research Fellow at the College of Computing and Data Science, Nanyang Technological University, Singapore, working with Prof. Shijian Lu. He received the B.S., M.S., and Ph.D. degrees from the School of Astronautics, Beihang University, Beijing, China, in 2019, 2022, and 2025, respectively, under the supervision of Prof. Zhenwei Shi and Prof. Zhengxia Zou. His research focuses on computer vision and remote sensing, with particular emphasis on foundation models, multimodal learning, and AI4Earth. He has authored or coauthored over 50 peer-reviewed papers in leading journals and conferences, including Proceedings of the IEEE, IEEE TPAMI, IEEE TGRS, and CVPR. His personal website is \url{https://chenkeyan.top}.
\end{IEEEbiography}

\begin{IEEEbiography}[{\includegraphics[width=1in,height=1.25in,clip,keepaspectratio]{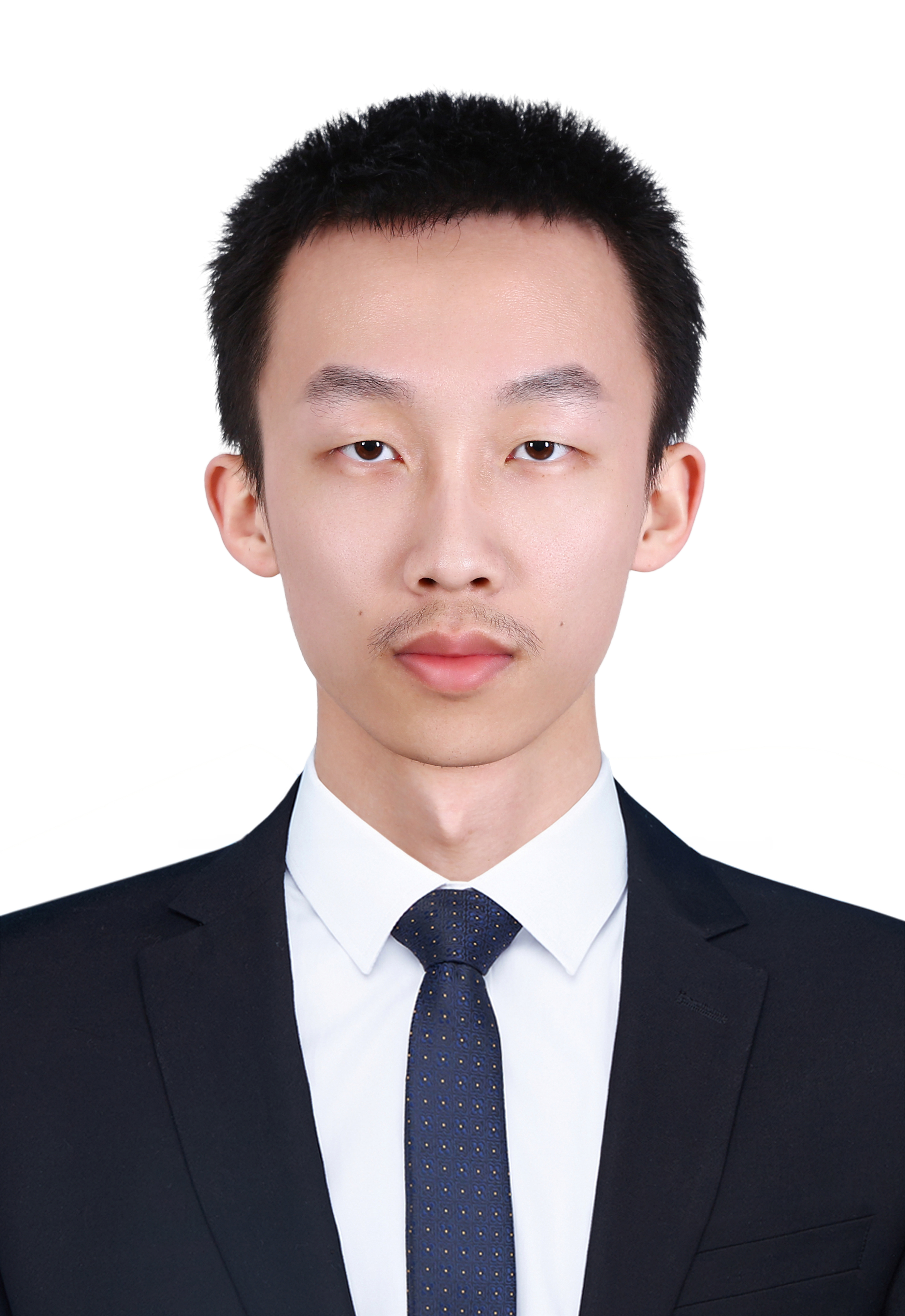}}]
{Qinzhe Yang}
is currently a student at the Shen Yuan Honors College, Beihang University, pursuing a Ph.D. degree through an eight-year integrated program in Future Aerospace Technology.

His research interests include image processing and deep learning, particularly object detection and instance segmentation in remote sensing.

\end{IEEEbiography}

\begin{IEEEbiography}[{\includegraphics[width=1in,height=1.25in,clip,keepaspectratio]{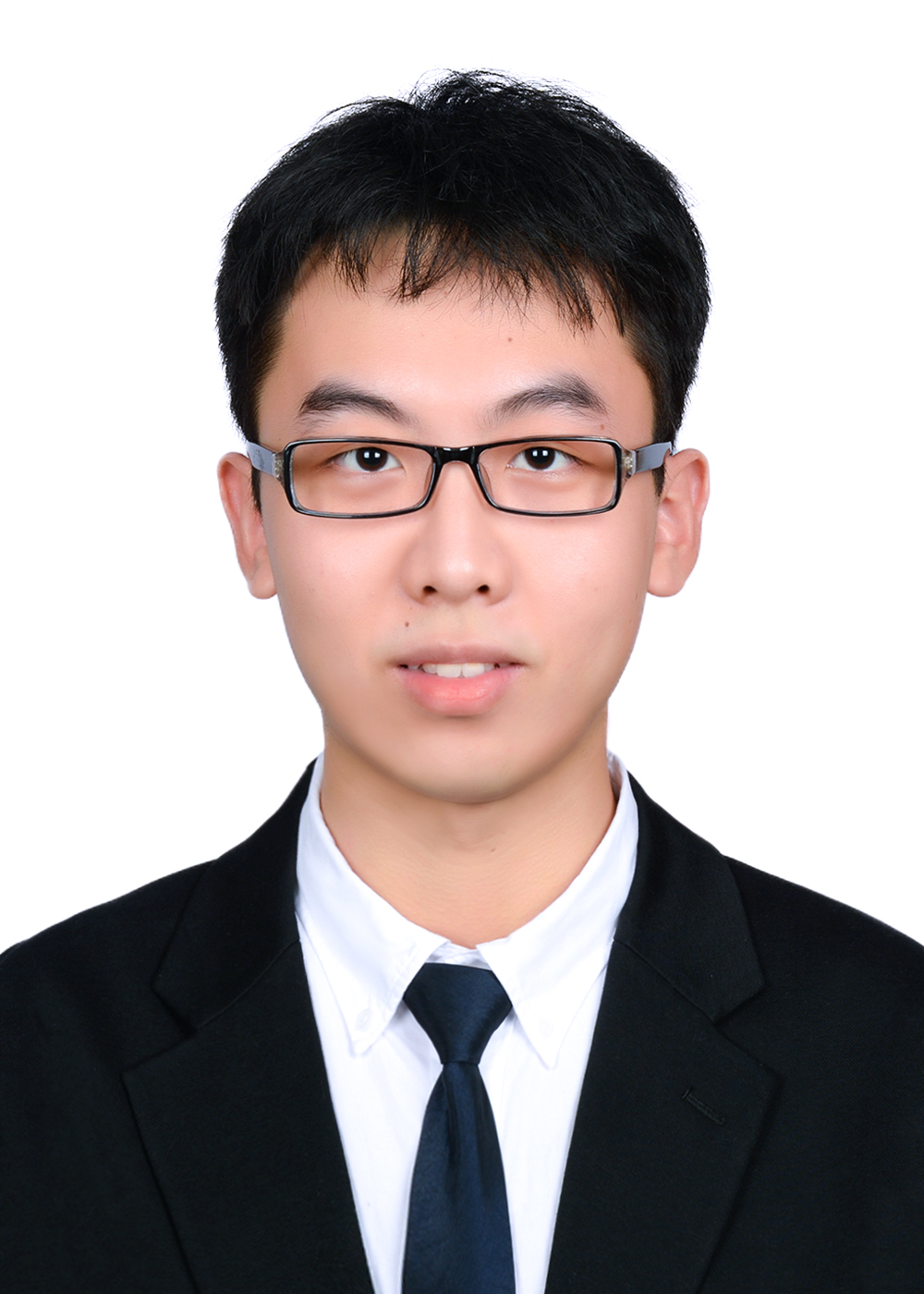}}]
{Bowen Chen}
received his B.S. degree from China University of Petroleum East China, Qingdao, Shandong, China, in 2022. He is currently working toward his doctor's degree in the School of Astronautics, Beihang University. 

His research interests include remote sensing image processing and computer vision.

\end{IEEEbiography}

\begin{IEEEbiography}[{\includegraphics[width=1in,height=1.25in,clip,keepaspectratio]{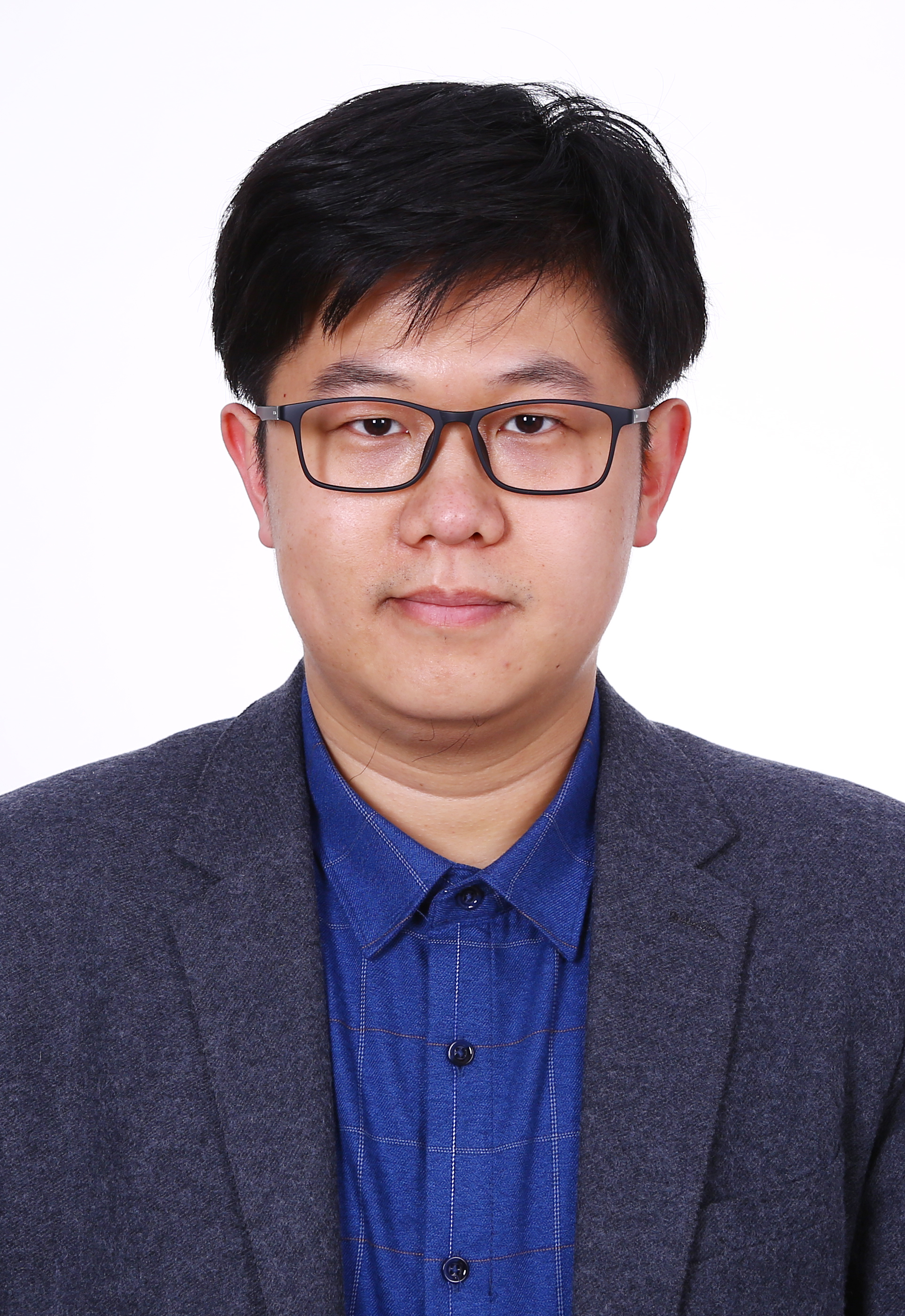}}]{Zhengxia Zou \text{(Senior Member, IEEE)}}
 received his BS degree and his Ph.D. degree from Beihang University in 2013 and 2018. He is currently a Professor at the Department of Aerospace Intelligent Science and Technology, School of Astronautics, Beihang University. During 2018-2021, he was a postdoc research fellow at the University of Michigan, Ann Arbor. His research interests include computer vision and related problems in remote sensing. He has published over 100 peer-reviewed papers in top-tier journals and conferences, including Proceedings of the IEEE, Nature Communications, IEEE Transactions on Pattern Analysis and Machine Intelligence, IEEE Transactions on Geoscience and Remote Sensing, and IEEE / CVF Computer Vision and Pattern Recognition. 
 
 Dr. Zou serves as the Associate Editor for IEEE Transactions on Image Processing and IEEE Transactions on Geoscience and Remote Sensing. His personal website is \url{https://zhengxiazou.github.io/}.
\end{IEEEbiography}

\begin{IEEEbiography}
[{\includegraphics[width=1in,height=1.25in,clip,keepaspectratio]{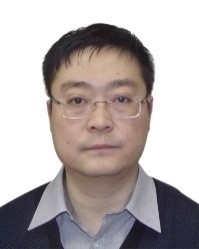}}]
{Zhenwei Shi \text{(Senior Member, IEEE)}}
is currently a Professor and Dean of the Department of Aerospace Intelligent Science and Technology, School of Astronautics, Beihang University. He has authored or co-authored over 300 scientific articles in refereed journals and proceedings, including the IEEE Transactions on Pattern Analysis and Machine Intelligence, the IEEE Transactions on Image Processing, the IEEE Transactions on Geoscience and Remote Sensing, the IEEE Conference on Computer Vision and Pattern Recognition (CVPR) and the IEEE International Conference on Computer Vision (ICCV). His current research interests include remote sensing image processing and analysis, computer vision, pattern recognition, and machine learning.

Prof. Shi serves as an Editor for IEEE Transactions on Geoscience and Remote Sensing, Pattern Recognition, ISPRS Journal of Photogrammetry and Remote Sensing, Infrared Physics and Technology, etc. His personal website is \url{http://levir.buaa.edu.cn/}.
\end{IEEEbiography}